# Oriented object detection in optical remote sensing images using deep learning: a survey

**Kun Wang[1,2] · Zi Wang[1,2] · Zhang Li[1,2] · Ang Su[1,2] · Xichao Teng[1,2] · Erting Pan[1,2] · Minhao Liu[3] · Qifeng Yu[1,2]**

Accepted: 2 May 2025

**Abstract**
Oriented object detection is a fundamental yet challenging task in remote sensing (RS), aiming to locate and classify objects with arbitrary orientations. Recent advancements in deep learning have significantly enhanced the capabilities of oriented object detection methods. Given the rapid development of this field, a comprehensive survey of the recent advances in oriented object detection is presented in this paper. Specifically, we begin by tracing the technical evolution from horizontal object detection to oriented object detection and highlighting the specific related challenges, including feature misalignment, spatial misalignment, oriented bounding box (OBB) regression problems, and common issues encountered in RS. Subsequently, we further categorize the existing methods into detection frameworks, OBB regression techniques, feature representation approaches, and solutions to common issues and provide an in-depth discussion of how these methods address the above challenges. In addition, we cover several publicly available datasets and evaluation protocols. Furthermore, we provide a comprehensive comparison and analysis involving the state-of-the-art methods. Toward the end of this paper, we identify several future directions for oriented object detection research.

## 1 Introduction

With the rapid advancement of remote sensing (RS) technologies, an increasing number of images with various resolutions and distinct spectra can be easily obtained by optical satellites or unmanned aerial vehicles (UAVs). Naturally, the research community has an imperative need to investigate a variety of advanced technologies for automatically and efficiently processing and analyzing massive numbers of RS images. As a pivotal foundation for automatically analyzing RS images, object detection is aimed at identifying objects belonging to predefined categories within the given images and regressing a precise localization for

Kun Wang, Zi Wang, and Zhang Li have contributed equally to this work.

Extended author information available on the last page of the article each object instance (Liu et al. 2020; Zou et al. 2023). Currently, object detection is a vital component in a broad range of RS applications, including intelligent monitoring (Zhao et al. 2018), precision agriculture (Osco et al. 2021), urban planning (Burochin et al. 2014), port management (Zhang et al. 2021), and military reconnaissance (Liu et al. 2022).

Objects in RS images typically exhibit arbitrary orientations due to their bird's-eye view (BEV) acquisition perspectives, making the general (horizontal) object detection methods inadequate. In contrast with general object detection, which involves performing object localization via a horizontal bounding box (HBB), oriented object detection (also called rotated object detection) employs an oriented bounding box (OBB) to tightly pack the oriented object, as shown in Fig. 1. This OBB can not only provide orientation information but also precisely locate the object. Consequently, oriented object detection has attracted considerable attention, especially within the past five years. Although many methods are available, a comprehensive survey specifically focused on oriented object detection is still lacking. Given the continued maturity and increasing concerns about this field, this paper seeks to present a thorough analysis of recent efforts and systematically summarize their achievements.

## 1.1 Comparisons with related surveys

Many prominent surveys have been published in the object detection field in recent years, as summarized in Table 1. Numerous notable surveys have concentrated on generic (horizontal) object detection, which aims to detect horizontal objects in natural scenarios (Liu et al. 2020; Wu et al. 2020; Zhao et al. 2019; Zou et al. 2023). These surveys covered various aspects, including deep-learning-based detection frameworks, training strategies, feature representation approaches, evaluation metrics, and typical applications.

Furthermore, several efforts have been devoted to specific categories, such as text detection (Ye and Doermann 2015) and pedestrian detection (Cao et al. 2022). Additionally, recent surveys have focused on object detection tasks conducted under specific conditions, including small object detection (Cheng et al. 2023), few-shot object detection (Pannone 2022), and weakly supervised object detection (Zhang et al. 2022).

Although a few surveys have analyzed and summarized the RS object detection domain, they frequently lack in-depth analyses of oriented object detection (Cheng and Han 2016; Li et al. 2020, 2021; Han et al. 2021; Wu et al. 2022). Zhang et al. (2023) classified the sub-

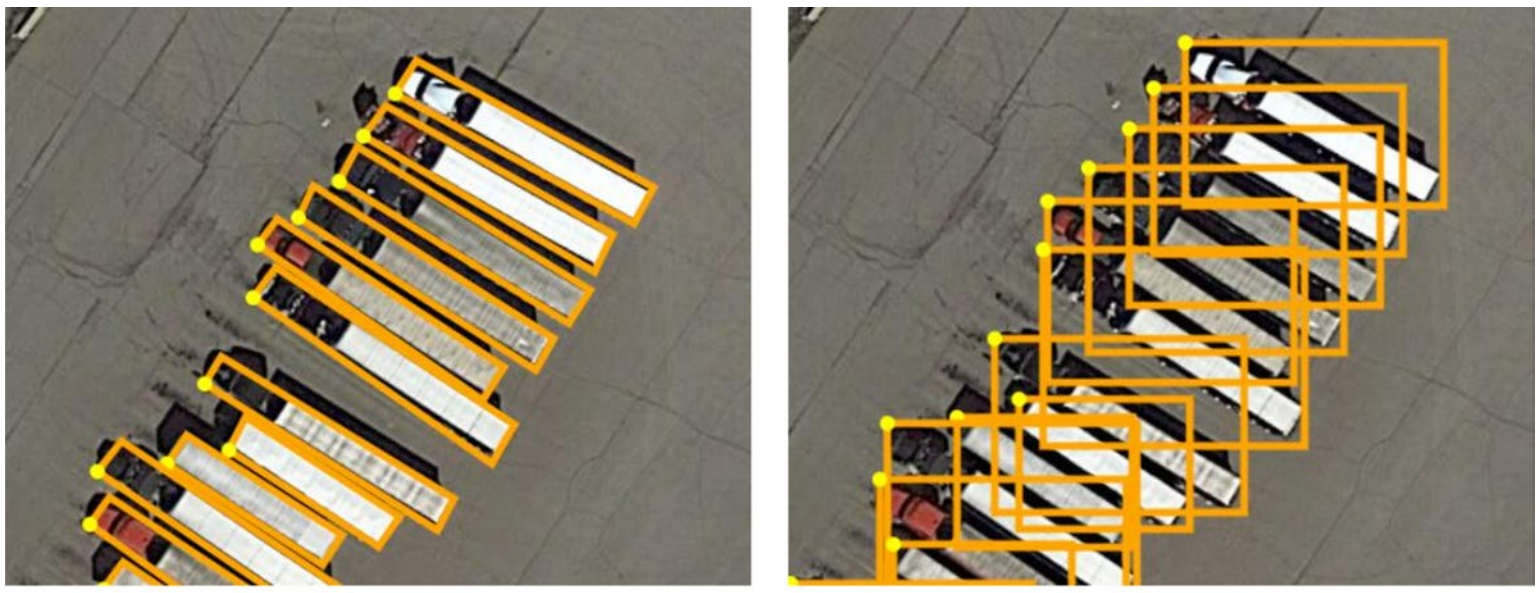

(a) OBB representation (b) HBB representation

**Fig. 1** Comparison between OBB and HBB (Xia et al. 2018; Ding et al. 2022). **a** OBB representation of objects. **b** is a failure case of the HBB representation, which brings high overlap compared to **a**

**Table 1** Summary of related object detection surveys in recent years

| 1.5px Survey title | Publication | Descriptions |
|---|---|---|
| 1.5px Deep learning for generic object detection: a survey (Liu et al. 2020) | IJCV 2020 | A comprehensive survey of the recent progress in generic object detection brought about by deep learning |
| Recent advances in deep learning for object detection (Wu et al. 2020) | Neucom 2020 | A survey focuses on deep learning in generic object detection from detection components, learning strategies, and applications |
| Object detection with deep learning: a review (Zhao et al. 2019) | TNNLS 2020 | A review on deep learning for generic object detection and other specific subtasks |
| Object detection in 20 years: a survey (Zou et al. 2023) | PROC 2023 | A survey focuses on object detection spanning over 20 years of history |
| Text detection and recognition in imagery: a survey (Ye and Doermann 2015) | TPAMI 2015 | A survey about methods, sub-problems, and special issues of text detection and recognition |
| From handcrafted to deep features for pedestrian detection: a survey (Cao et al. 2022) | TPAMI 2022 | A survey on recent deep features based methods in pedestrian detection |
| Towards large-scale small object detection: survey and benchmarks (Cheng et al. 2023) | TPAMI 2023 | A survey of small object detection and two large-scale small object detection benchmarks under driving scenario and aerial scene |
| Few-shot object detection: a survey (Pannone 2022) | ACM 2022 | A survey on few-shot object detection through data augmentation, transfer learning, distance metric learning, and meta-learning |
| A survey on object detection in optical remote sensing images (Cheng and Han 2016) | ISPRS 2016 | A review on traditional object detection methods in RS images |
| Object detection in optical remote sensing images: a survey and a new benchmark (Li et al. 2020) | ISPRS 2020 | A review on deep learning based horizontal object detection in RS, and a large-scale, publicly available benchmark for RS object detection |
| Remote sensing object detection meets deep learning: a meta-review of challenges and advances (Zhang et al. 2023) | GRSM 2023 | A survey on challenges and advances in RS object detection, including multiscale object detection, rotated object detection, weak object detection, tiny object detection, and object detection with limited supervision |
| Ship detection and classification from optical remote sensing images: a survey (Li et al. 2021) | CJA 2021 | A survey of RS ship detection schemes from 1978 to 2020 |
| Methods for small, weak object detection in optical high-resolution remote sensing images: a survey of advances and challenges (Han et al. 2021) | GRSM 2021 | A survey of challenges and recent advances for RS small, weak object detection |
| Deep learning for unmanned aerial vehicle-based object detection and tracking: a survey (Wu et al. 2022) | GRSM 2022 | A survey on deep learning approaches in UAV object detection and tracking from static object detection, video object detection, and multiple object detection |
| A comprehensive survey of oriented object detection in remote sensing images (Wen et al. 2023) | ESWA 2023 | A survey on oriented object detection, including rotation invariance, anchor-free mechanism, and loss function. |

*Top* generic object detection. *Middle* object detection focusing on specific tasks. *Bottom* RS object detection categories belonging to RS object detection as oriented object detection, providing only a brief introduction to OBB representation and rotation-insensitive feature learning methods. Wen et al. (2023) focused only on describing the details of previously developed oriented object detection methods.

Unlike previously presented object detection surveys, which focused on general object detection methods (Liu et al. 2020; Wu et al. 2020; Zhao et al. 2019; Zou et al. 2023), other related fields (Ye and Doermann 2015; Cao et al. 2022; Cheng et al. 2023; Pannone 2022; Zhang et al. 2022), horizontal RS object detection (Cheng and Han 2016; Li et al. 2020, 2021; Han et al. 2021; Wu et al. 2022), or limited numbers of oriented object detection models (Zhang et al. 2023; Wen et al. 2023), this work systematically and comprehensively reviews the recent advances in the field. In particular, relative to the existing surveys concerning oriented object detection (Zhang et al. 2023; Wen et al. 2023), our survey provides a deeper, more comprehensive dive into this field; offers a better taxonomy of the literature; and presents discussions regarding the current challenges, comparisons, and future research directions. It involves in-depth analyses of various aspects, many of which, to our knowledge, have never been discussed in oriented object detection surveys. In particular, we review the technical evolution from horizontal to oriented object detection and summarize the main challenges. We systematically summarize and discuss the recent advancements developed under the proposed taxonomies (including detection frameworks, OBB regression techniques, feature representation approaches, and solutions to common issues). We provide a comprehensive comparison among the state-of-the-art methods on typical datasets, along with an in-depth analysis of the pros and cons of these methods.

## 1.2 Scope

Fig. 2 shows the increasing number of publications related to "oriented object detection" or "rotated object detection" over the past decade. In particular, in the last five years, explosive growth has been observed in the number of papers published on deep-learning-based oriented object detection methods, rendering it impractical to review all of them. Conse-quently, it is necessary to establish selection criteria to limit our focus to the influential papers published in top journals and conferences. Owing to these constraints, we extend our sincere apologies to authors whose works are not included in this paper. Notably, we restrict our attention to oriented object detection methods for single images. Nevertheless, for completeness and improved readability, some well-known works on horizontal object detection are also included.

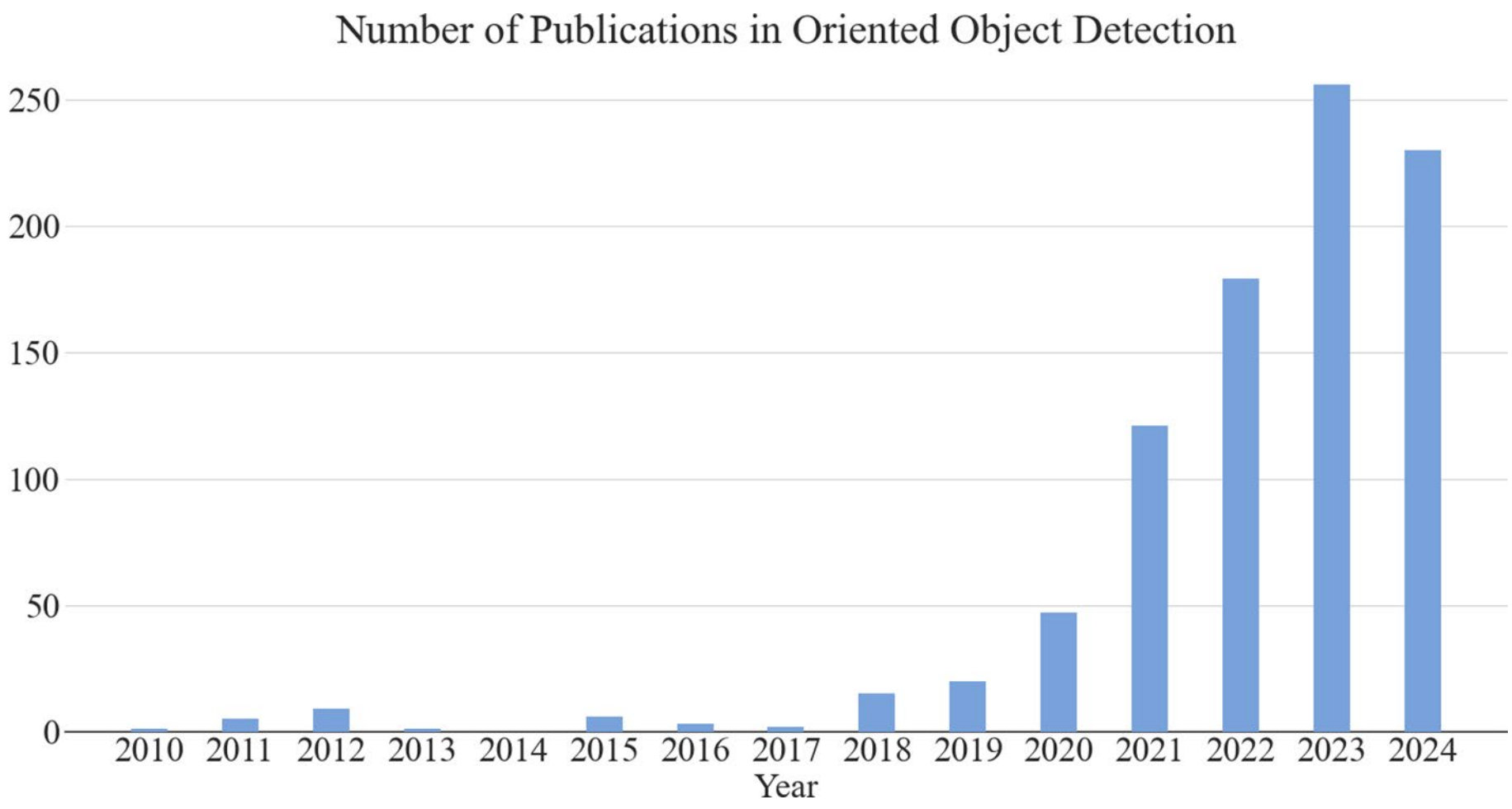

**Fig. 2** Increasing number of publications in oriented object detection from 2010 to 2024

### 1.3 Contributions

Our contributions are manifested in four aspects.

(1) *A comprehensive review of the technical evolution from horizontal object detection to oriented object detection* On the basis of the characteristics of RS images and the current object detection models, we categorize the main challenges involved in oriented object detection into four main parts: feature misalignment, spatial misalignment, OBB regression problems, and common issues encountered in RS.

(2) *A thorough taxonomy of the existing oriented object detection methods* To help researchers gain a deeper understanding of the key features of oriented object detection methods, we categorize and summarize the existing oriented object detection methods according to detection frameworks, OBB regression techniques, feature representation approaches, and solutions to common issues.

(3) *A comprehensive comparison among the state-of-the-art methods* We provide a comprehensive comparison among the state-of-the-art methods on typical datasets, along with an in-depth analysis of the advantages and disatvantages of these methods. This analysis aims to offer valuable insights into the efficacy and applicability of these methods in terms of addressing the main challenges associated with oriented object detection.

(4) *Overview of the open issues and future research directions* We thoroughly examine several essential issues, shedding light on potential directions for future research, i.e., lightweight methods, scenario-specific datasets, multimodal datasets, and large-scale datasets, as well as multimodal large models.

The structure of this paper is organized as follows. We first introduce the development trend from horizontal object detection to oriented object detection and highlight the major related challenges in Sect. 2. We review the existing deep neural network (DNN)-based detection frameworks in Sect. 3. Furthermore, we discuss the OBB regression techniques and feature representation approaches in Sects. 4 and 5, respectively. In addition, we summarize the solutions to other common issues encountered in RS scenarios in Sect. 6. After an overview of the commonly used datasets is provided in Sect. 7, we analyze and compare the state-of-the-art methods in Sect. 8. Finally, we conclude our work and discuss future directions for oriented object detection research in Sect. 9.

## 2 From horizontal object detection to oriented object detection

The early object detection methods relied on handcrafted descriptors (Lowe 2004; Dalal and Triggs 2005; Fei-Fei and Perona 2005; Wright et al. 2009) and machine learning algorithms (Cortes and Vapnik 1995; Blaschke 2010; Leitloff et al. 2010; Blaschke et al. 2014). These methods often have limited performance because of their weak feature representa-tions. Although such approaches lag far behind deep learning methods in terms of accuracy, their instructive insights still have profound impacts on modern detectors, e.g., sliding windows (Viola and Jones 2001, 2004), hard negative mining, and bounding box regression (Felzenszwalb et al. 2008, 2010). Readers interested in these early object detection methods are referred to a recent survey (Cheng and Han 2016) that provided an in-depth analysis of the classic RS object detection methods.

The world has witnessed impressive progress in computer vision with the advances achieved in deep neural networks (DNNs) since 2012 (Hinton and Salakhutdinov 2006; LeCun et al. 2015; Chen et al. 2018; He et al. 2016; Krizhevsky et al. 2012, 2017). Owing to the persistent increases in computing resources, DNNs can learn high-level patterns from large-scale datasets in an end-to-end manner. The pioneering studies bring a little glimmer to the object detection field, especially because the performance of handcrafted-feature-based detectors reached a plateau after 2010. Since then, a growing number of DNN-based detectors have emerged and have dominated the state-of-the-art methods due to their powerful feature representation capabilities.

The early research in the deep learning era was primarily concerned with designing horizontal object detectors (Girshick et al. 2014; Girshick 2015; Ren et al. 2015, 2017; Liu et al. 2016; Lin et al. 2017, 2020; Redmon et al. 2016; Redmon and Farhadi 2017; Hei and Jia 2020; Duan et al. 2019; Zhou et al. 2019; Yang et al. 2019) for natural scene images taken from a horizontal perspective. Naturally, as horizontal object detectors rapidly evolve, e.g., the region-based convolutional neural network (RCNN) series (Girshick et al. 2014; Girshick 2015; Ren et al. 2015, 2017), You Only Look Once (YOLO) series (Redmon et al. 2016; Redmon and Farhadi 2017), and RetinaNet (Lin et al. 2017, 2020), numerous studies are harnessing their immense potential in RS scenarios. A growing number of efforts have focused on refining network structures and crafting innovative data augmentation techniques, all of which have been aimed at addressing the core challenges encountered in RS object detection scenarios, including scale variations (Liang et al. 2020; Ye et al. 2022; Liu et al. 2022; Khan et al. 2022; Li et al. 2023; Wang et al. 2025), complex backgrounds (Lu et al. 2021; Huang et al. 2022; Ma et al. 2022; Zhou et al. 2023), and weak feature responses (Tian et al. 2022; Wu et al. 2022; Zhuang et al. 2024).

Nevertheless, RS images are typically captured from a BEV, leading to objects appearing with arbitrary orientations. Hence, directly applying horizontal object detectors to RS images may lead to the following problems. (1) The intersection-over-union (IoU) between one HBB and the adjacent HBBs can be very large in densely arranged scenarios, especially for objects with extremely large aspect ratios, as illustrated in Fig. 1b. Thus, the nonmaximum suppression (NMS) technique tends to result in missed detections. (2) HBBs are inclined to contain background areas, whereas OBBs can tightly enclose objects, thereby achieving more precise localization effects, as shown in Fig. 1a. Given the above predicament concerning HBBs, OBBs are considered more appropriate for RS object detection tasks.

With the remarkable development of detection frameworks (Ren et al. 2015, 2017; Lin et al. 2017, 2020; Redmon et al. 2016; Carion et al. 2020), backbone networks (He et al. 2016; Liu et al. 2021), and robust feature representations (Liu et al. 2022; Dosovitskiy et al. 2021), the field of object detection has achieved dramatic breakthroughs. Naturally, an intuitive strategy for designing oriented object detectors is to modify the representative horizontal object detectors by predicting additional parameters to represent OBBs (Zhou et al.

2022). However, such a straightforward strategy is plagued by several additional challenges, including feature misalignment, spatial misalignment, and OBB regression problems.

(1) *Feature misalignment* The prevailing generic object detectors typically contain a feature extraction network followed by a detection head, where the latter leverages the feature representations generated by the former to make decisions. However, these feature representations are generally extracted via axis-aligned convolutions, thereby exposing non-negligible misalignment issues w.r.t. the oriented objects, as shown in Fig. 3a. This type of misaligned feature representations degrades the performance of oriented object detectors because of their lack of rotational information, making the detectors struggle to identify objects and regress precise OBBs.

(2) *Spatial misalignment* In addition to feature misalignment, the widely used anchor-based detection methods also struggle with spatial misalignment. Generic anchor-based detectors typically use horizontal anchors as priors, thereby having limited IoUs with oriented objects, especially for objects with extremely large aspect ratios, as shown in Fig. 4a. This poses a significant challenge to generic label assignment strategies (Ren et al. 2015, 2017), which assign positive or negative samples depending on their IoUs. Thus, the naïve anchor generation mechanism is likely unable to provide sufficient positive samples during the training process.

(3) *OBB regression problems* The current detectors commonly use the regression paradigm to represent the locations of objects, which has been proven to be an effective approach and has yielded dramatic achievements. The most commonly used representation methods for OBBs include $\theta$-based and quadrilateral representations. However, the former suffers from the periodicity of angles (PoA), causing angular boundary discontinuities (Yang et al. 2021, 2022; Qian et al. 2021, 2022). Specifically, a small angle difference may cause a large loss change when the angular value approaches the angular boundary range. On the other

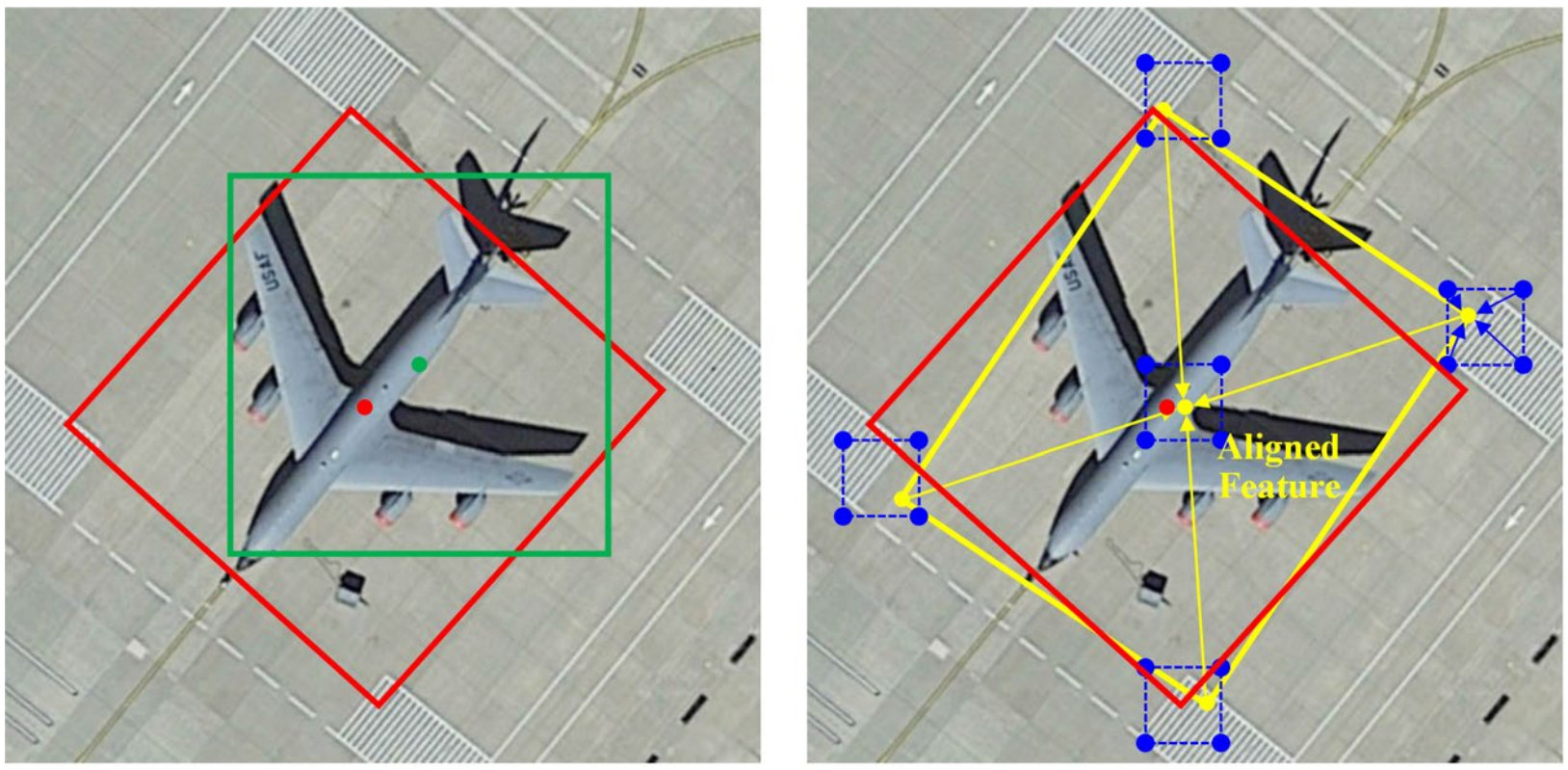

(a) Feature misalignment

(b) Feature alignment of $R^3$Det (Yang et al., 2021)

**Fig. 3** Illustration of feature misalignment and feature alignment. **a** The misalignment between oriented objects and the axis-aligned feature representations of anchor. **b** A example of feature alignment proposed by $R^3$Det (Yang et al. 2021), which align the feature representations by integrating the features according to the five refined points of the predicted obb. The red, green, and blue boxes represent the ground truth (GT), anchor, and predicted obb, respectively. The blue and yellow points denote anchor points and refined feature points, respectively. The blue and yellow arrows denote feature interpolation and feature alignment operation, respectively

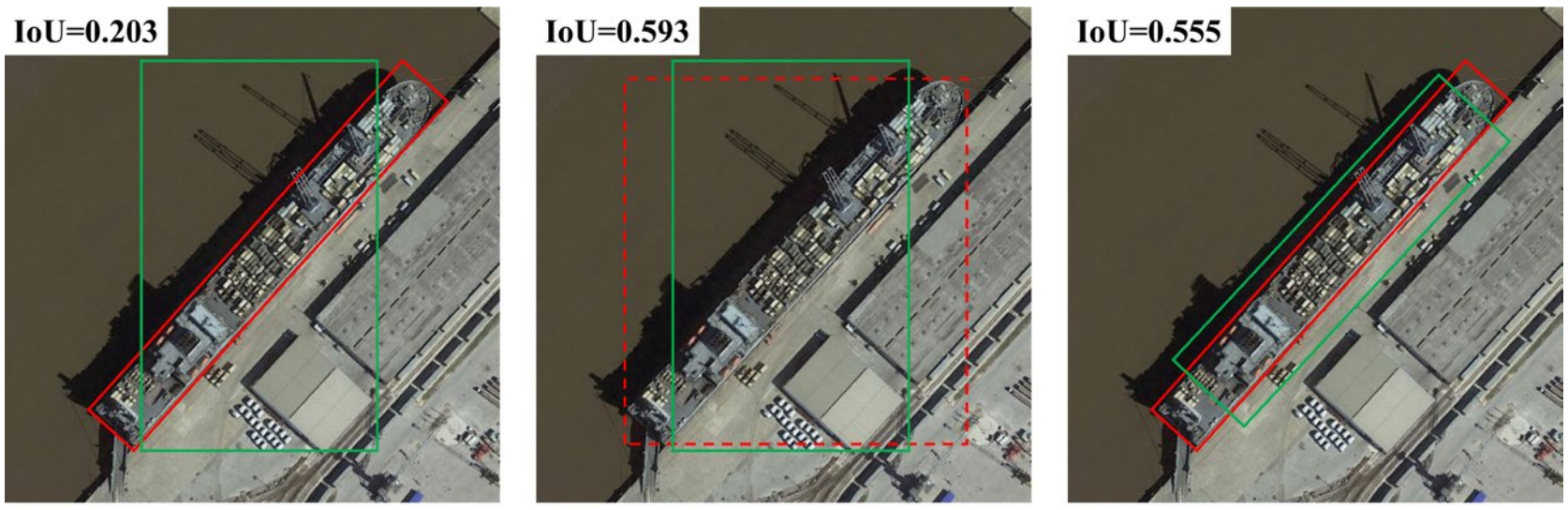

(a) (b) (c)

**Fig. 4** Illustration of spatial misalignment. **a** The IoU between horizontal anchor and oriented object is very small, causing spatial misalignment. **b**, **c** Calculating the IoU between either horizontal anchor and horizontal bounding rectangle of object, or rotated anchor and oriented object, can alleviate the spatial misalignment. The red and green boxes represent the GT and anchor, respectively

hand, the latter faces challenges related to vertex ordering because an inappropriate vertex sorting process may cause inconsistencies between the vertex sequences of the predicted OBB and the ground truth (GT). Overall, both the PoA and vertex ordering problems can seriously confuse the networks, leading to training instability. For more details, please refer to Sects. A and B of the Appendix.

To address the above dilemmas, various works have been conducted and have achieved notable advancements. Several methods construct well-designed *detection frameworks* by devising rotated proposal generation networks (Sect. 3.1) or refined heads (Sect. 3.2) to remedy feature misalignment. To address spatial misalignment, researchers have focused on improving their assignment schemes (Sect. 3.1) and adopting anchor-free mechanisms (Sect. 3.3). For OBB regression problems, several efforts have been made to design effective *OBB regression technologies* through the development of new loss functions (Sect. 4.1) and OBB representation schemes (Sect. 4.2). In addition, high-quality *feature representations* are crucial for performing object detection; hence, much effort is concentrated on designing networks to produce better feature representations, including rotation-invariant feature representations (Sect. 5.1) and advanced feature representations (Sect. 5.2). We also cover the solutions for addressing several *common issues* (Sect. 6) encountered in RS scenarios. Figure 5 shows the taxonomy of the oriented object detection methods examined in this survey.

## 3 Detection frameworks

Object detection methods can be categorized into two primary groups: two-stage and one-stage detection (Liu et al. 2020; Zou et al. 2023). The former work in a coarse-to-fine paradigm, whereas the latter accomplish classification and regression in one step, thereby exhibiting high efficiency but poor accuracy. In contrast with the aforementioned two categories, which rely on anchor mechanisms, anchor-free methods directly detect objects without the need for predefined anchors. In addition, a series of DETR-based methods have recently emerged; such methods regard the detection process as a set prediction task, thereby effectively eliminating several handcrafted components, e.g., NMS and anchor mechanisms. Considering that each category has advantages and disadvantages, we divide the representa-tive oriented object detectors into four categories: two-stage, one-stage, anchor-free, and DETR-based. Several key methods are presented in Fig. 6. Next, we concisely review how each category addresses the feature misalignment and spatial misalignment via deliberate framework designs.

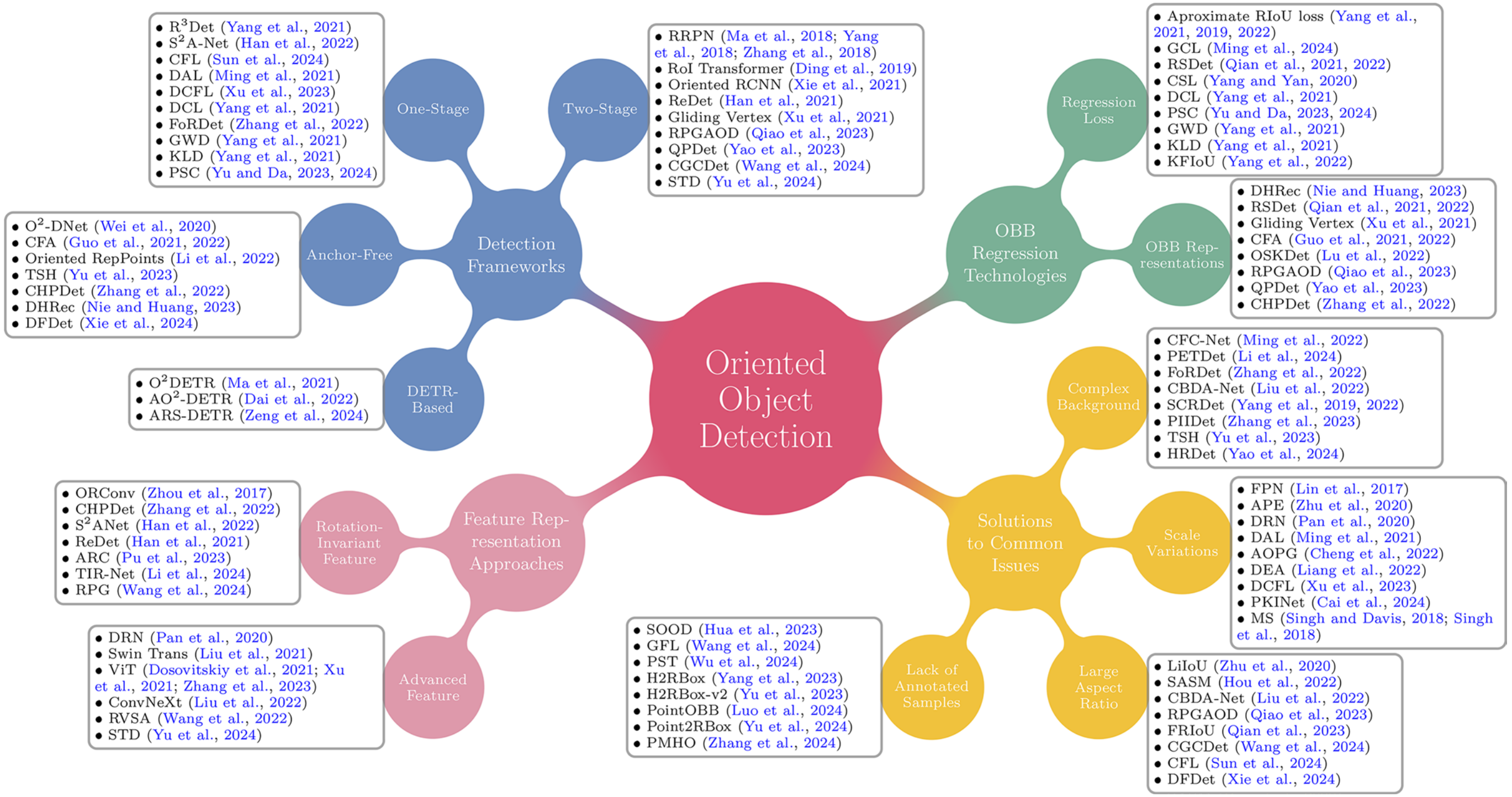

**Fig. 5** Structured taxonomy of the deep learning based oriented object detection methods in this survey

### 3.1 Two-stage detectors

Among the numerous prominent two-stage detectors (Ren et al. 2015, 2017; Lin et al. 2017; Cai and Vasconcelos 2018; He et al. 2020; Qiao et al. 2021), the Faster RCNN (Ren et al. 2015, 2017) armed with feature pyramid networks (FPN) (Lin et al. 2017) is commonly used as a benchmark owing to its exceptional accuracy and efficient design. As depicted in Fig. 7a, its workflow consists of the following pipeline: a feature extraction module, a region proposal network (RPN), and an RCNN. In the first stage, a sparse set of high-quality region proposals that can potentially contain objects is generated via the RPN (Chavali et al. 2016; Hosang et al. 2016). During the second stage, the region features of each proposal are extracted and then used for classification and refined regression processes via the RCNN. Finally, several post-processing operations, such as NMS, are leveraged to finalize the detection results (omitted in Fig. 7a). The oriented version of this method, termed Rotated Faster RCNN or Faster RCNN OBB, predicts the orientation of each object by adding an extra channel in its regression branch.

However, the naïve RPN only generates horizontal region proposals as regions of interest (RoIs), as shown in Fig. 8a. Apart from the feature misalignment caused by axis-aligned convolutions, another factor that may impair the performance is the feature misalignment between the horizontal RoI (HRoI) and the OBB, as shown in Fig. 9a. The feature misalignment significantly harms the feature representations, making the utilized detector struggle to identify objects and regress precise OBBs yet inspiring successive innovations.

To cope with feature misalignment, various efforts have been dedicated to generating rotated proposals and then adopting rotated RoI (RRoI) operators to extract spatially aligned features, as shown in Fig. 9b. The RRPN (Ma et al. 2018; Yang et al. 2018; Zhang et al. 2018) incorporates rotated anchors to accommodate objects with various orientations. In addition to scales and aspect ratios, different orientation parameters can be added to further generate additional rotated anchors, as shown in Fig. 8b. This method can alleviate spatial misalignment (as shown in Fig. 4c), thereby achieving better performance in terms of recall. However, redundant rotated anchors lead to significantly increased computational costs and memory footprints.

To reduce the number of rotated anchors, the RoI Transformer (Ding et al. 2019) retains the naïve RPN structure to alleviate spatial misalignment (as shown in Fig. 4b) and then introduces a lightweight RoI learner module. As shown in Fig. 8c, the RRoI learner converts HRoIs directly into RRoIs, generating precise RRoIs without enormous numbers of rotated anchors, thus increasing the efficiency and accuracy. However, the added complexity of the RoI learner, which includes an extra RoI operator and a regression stage, makes the network less efficient.

Consequently, Xie et al. (2021) designed a simpler structure, Oriented RCNN, to directly generate high-quality RRoIs from horizontal anchors, as shown in Fig. 8d. This lightweight module benefits from the proposed midpoint offset representation, which includes the corresponding external HBB and the offsets of vertices w.r.t. the midpoints of the external

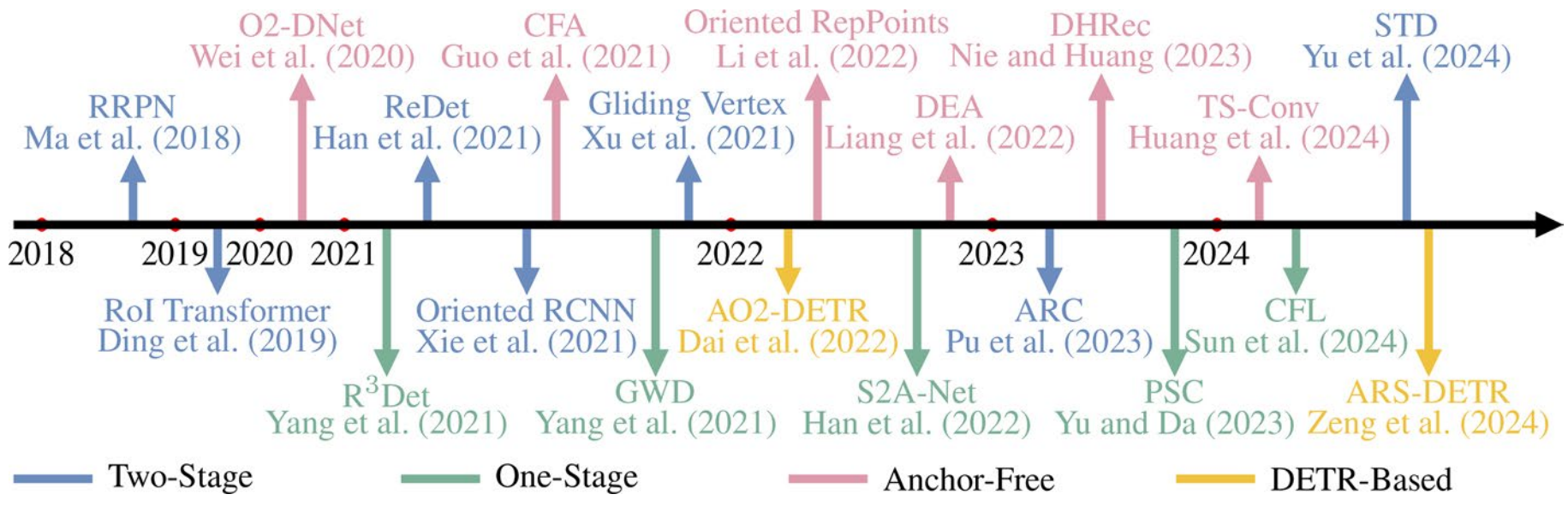

**Fig. 6** Chronological overview of the representative oriented object detection frameworks

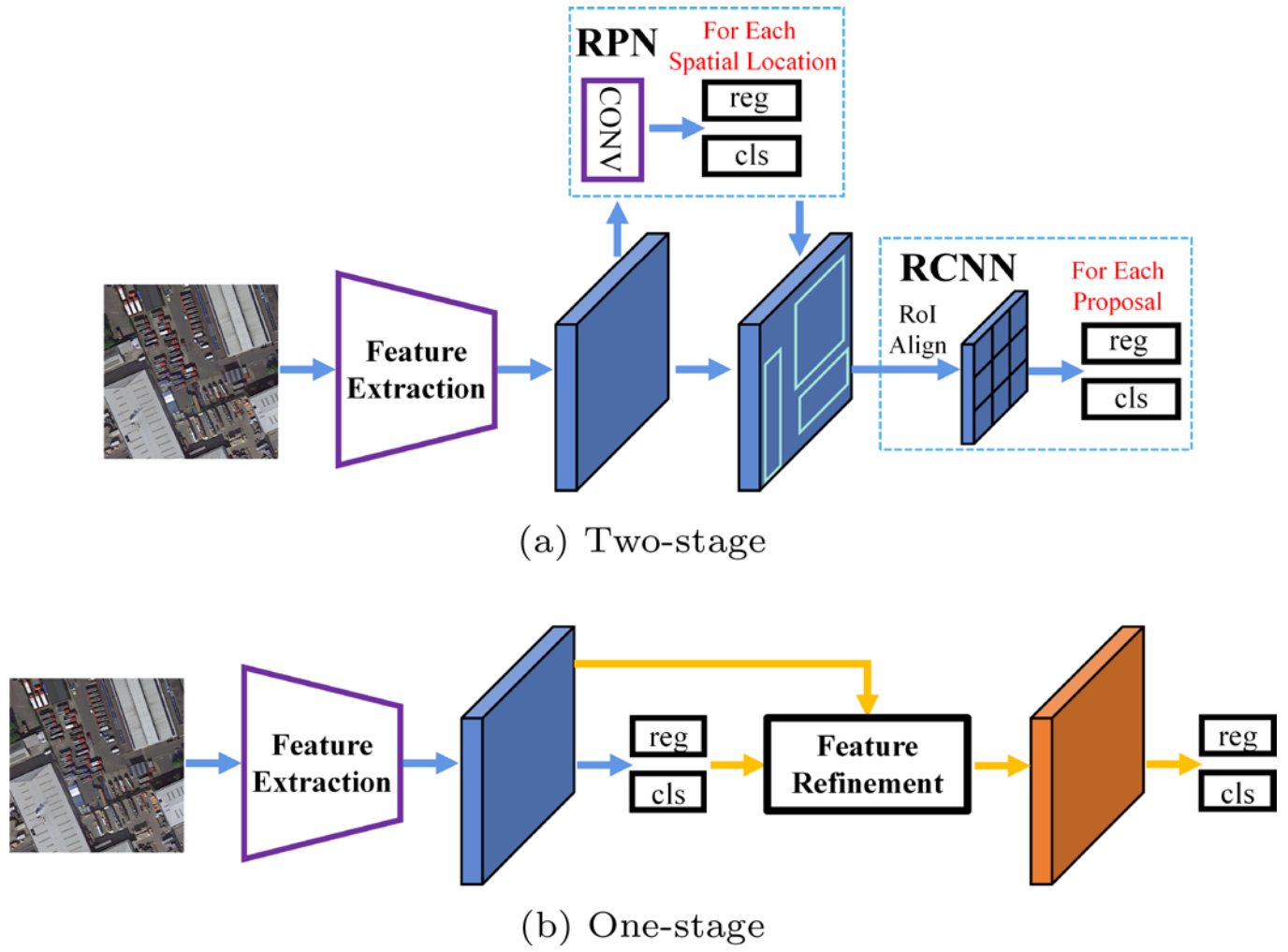

**Fig. 7** The basic architecture of two-stage and one-stage detectors. **a** Two-stage detectors first utilize RPN to predict a set of proposals, then extract corresponding region features for classification and refined regression. **b** One-stage detectors predict the class probabilities and locations for each spatial location. Most of them add a refined stage to alleviate the feature misalignment. The blue arrows denote the workflow of RetinaNet (Lin et al. 2017, 2020), while the orange arrows denote the workflow of refined stage

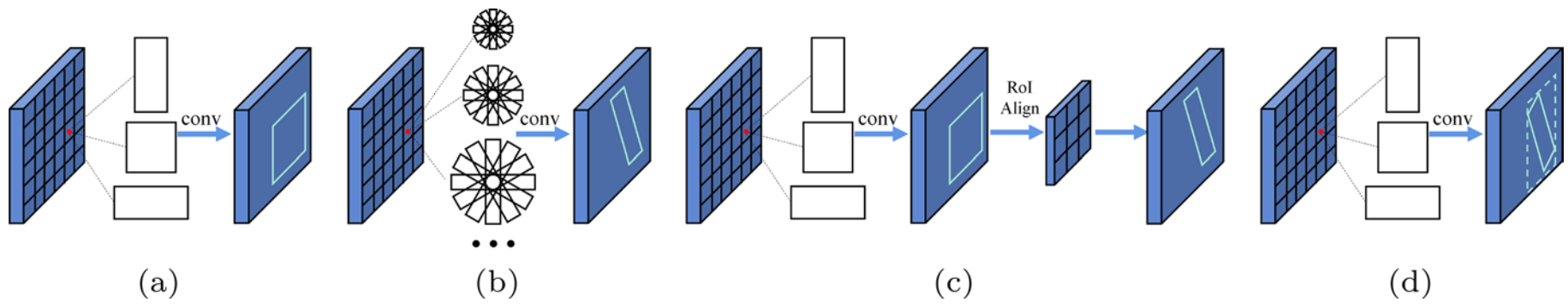

**Fig. 8** The comparisons of different strategies for proposal generation. **a** RPN only generates horizontal proposals (Ren et al. 2015, 2017). **b** RRPN densely places rotated anchors with different scales, ratios, and angles (Ma et al. 2018; Yang et al. 2018; Zhang et al. 2018). **c** RoI Transformer generates rotated proposal from horizontal RoI via RPN, RoI Alignment, and OBB regression (Ding et al. 2019). **d** Oriented RCNN can generate high-quality rotated proposals using a lightweight module (Xie et al. 2021)

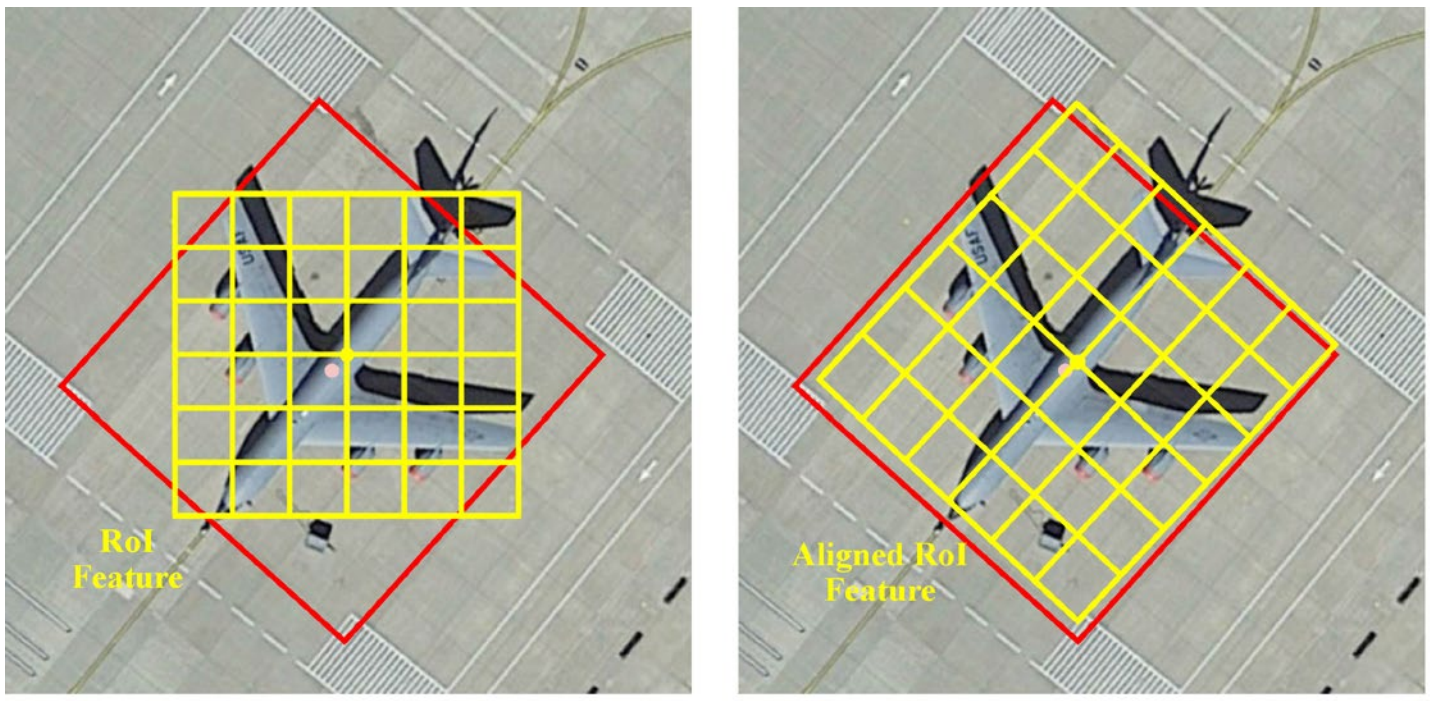

(a) Feature misalignment (b) Feature alignment of rotated proposal

**Fig. 9** Illustration of feature misalignment in two-stage detectors. **a** The feature misalignment between oriented object and horizontal region proposal. **b** Detectors can extract aligned features from rotated region proposal

HBB. This representation scheme maintains horizontal regression mechanisms, ensuring a more stable training process relative to regressing OBBs from horizontal anchors. Benefiting from the design of the oriented RPN and midpoint offset representation, the Oriented RCNN can achieve competitive accuracy w.r.t. advanced two-stage detectors and reach approximate efficiency in comparison with one-stage detectors.

The excellent designs of the RoI Transformer and Oriented RCNN address the spatial misalignment and feature misalignment faced by two-stage detectors, laying the foundation for subsequent research in this field. Numerous two-stage detectors have since adopted the RoI Transformer or Oriented RCNN as their baseline framework, harnessing exceptional feature representations (e.g., ReDet (Han et al. 2021), RVSA (Wang et al. 2022), ARC (Pu et al. 2023), and STD (Yu et al. 2024)) or crafting superior OBB representations (e.g., Gliding Vertex (Xu et al. 2021), RPGAOD (Qiao et al. 2023), and QPDet (Yao et al. 2023)) and loss functions (e.g., FRIoU (Qian et al. 2023), CGCDet (Wang et al. 2024), and GCL (Ming et al. 2024)) to increase their performance. These methods will be discussed in detail in the subsequent section.

## 3.2 One-stage detectors

As illustrated in Fig. 7b, one-stage detectors first extract multilevel feature maps and then predict the class probabilities and locations for each anchor per spatial location. Owing to the absence of RPN and RoI operators, one-stage detectors encounter more severe feature misalignment than two-stage detectors do. Thus, a series of one-stage algorithms, such as $R^3$Det (Yang et al. 2021) and $S^2$A-Net (Han et al. 2022), have been developed to alleviate this dilemma.

$R^3$Det (Yang et al. 2021) adopts a feature refinement module (FRM) to align features. First, $R^3$Det transforms its horizontal anchors into rotated anchors, which can provide more accurate positional and oriented information. Then, the FRM employs pixelwise feature interpolation to integrate features derived from the five locations (i.e., one center and four corners) of the corresponding refined rotated anchors, as shown in Fig. 3b. Similarly, $S^2$

A-Net (Han et al. 2022) aligns features via alignment convolution (AlignConv), which is a variant of deformable convolution (Dai et al. 2017). The offset field of AlignConv is inferred from the guidance of rotated anchors. Both the FRM and AlignConv operate in a coarse-to-fine manner but differ significantly from the RRoI operator. Notably, they follow a full convolution structure with fewer sampling points, resulting in increased efficiency.

Based on $S^2$A-Net, CFL (Sun et al. 2024) introduces a spatial transform selection (STS) strategy and a critical feature sampling (CFS) module. STS dynamically assigns labels by calculating IoU thresholds on the basis of aspect ratios, angle differences, and the initial IoU threshold determined by ATSS (Zhang et al. 2020). The adaptable IoU threshold controls the number of samples assigned to easy objects while ensuring that sufficient positive samples are provided for hard objects with large aspect ratios and angle differences. CFS incorporates a deformable convolution in which the sampling positions are derived from the initial detection results (center point, vertices, and midpoints) combined with a learnable offset field.

Both $R^3$Det and $S^2$A-Net utilize an extra refined head for feature alignment, making them popular baseline choices as one-stage detectors. These methods further solve the spatial misalignment and OBB regression problems via innovative sample assignment strategies (e.g., DAL (Ming et al. 2021) and DCFL (Xu et al. 2023)) or regression losses (e.g., DCL (Yang et al. 2021), FoRDet (Zhang et al. 2022), GWD (Yang et al. 2021), KLD (Yang et al. 2021), and PSC (Yu and Da 2023, 2024)). In the following section, we provide a detailed introduction to these methods.

### 3.3 Anchor-free detectors

The above two categories follow the anchor paradigm, which suffers from spatial misalignment between horizontal anchors and OBBs. To address the above issues, numerous anchor-free methods have been developed to detect objects without relying on preset anchors. These methods eliminate anchor-related hyperparameters, showing potential for generalization to wide-ranging applications (Zhang et al. 2020). According to their representations of OBBs, anchor-free methods can be divided into keypoint-based methods and center-based methods.

Keypoint-based methods first locate a set of adaptive or self-constrained keypoints and then circumscribe the spatial extent of an object, as shown in Fig. 10a. For example, $O^2$-DNet (Wei et al. 2020) first locates the midpoints of four sides of the OBB by regressing the offsets from the center point. The two sets of opposite midpoints are subsequently connected to form two mutually perpendicular midlines, which are decoded to represent the OBB. In addition, a self-supervision loss constrains the perpendicular relationship between two middle lines and the collinear relationship between the center point and the two opposite midpoints. Following RepPoints (Yang et al. 2019), CFA (Guo et al. 2021, 2022) uses deformable convolution (Dai et al. 2017) to generate a convex hull for each oriented object. The convex hull, which is represented by irregular sample points, is refined using a convex IoU (CIoU) loss. To alleviate the feature aliasing problem between densely packed objects, convex hull set splitting and feature antialiasing strategies are designed to refine the convex hulls and adaptively optimize the feature assignment process.

Furthermore, to predict high-quality oriented reppoints, Oriented RepPoints (Li et al. 2022) employs an adaptive point assessment and assignment (APAA) scheme to measure the quality of the reppoints. The APAA process assesses reppoints across four dimensions (classification, localization, orientation alignment, and pointwise correlation) to select high-quality points without imposing an additional computational burden during inference. Subsequently, Yu et al. (2023) proposed a dynamic information aggregation (DIA) module based on a multihead self-attention mechanism (Vaswani et al. 2017). By mining the relationships between reppoints, DIA not only helps obtain more accurate reppoint positions but also enriches the feature representations, thereby further increasing the model localization accuracy.

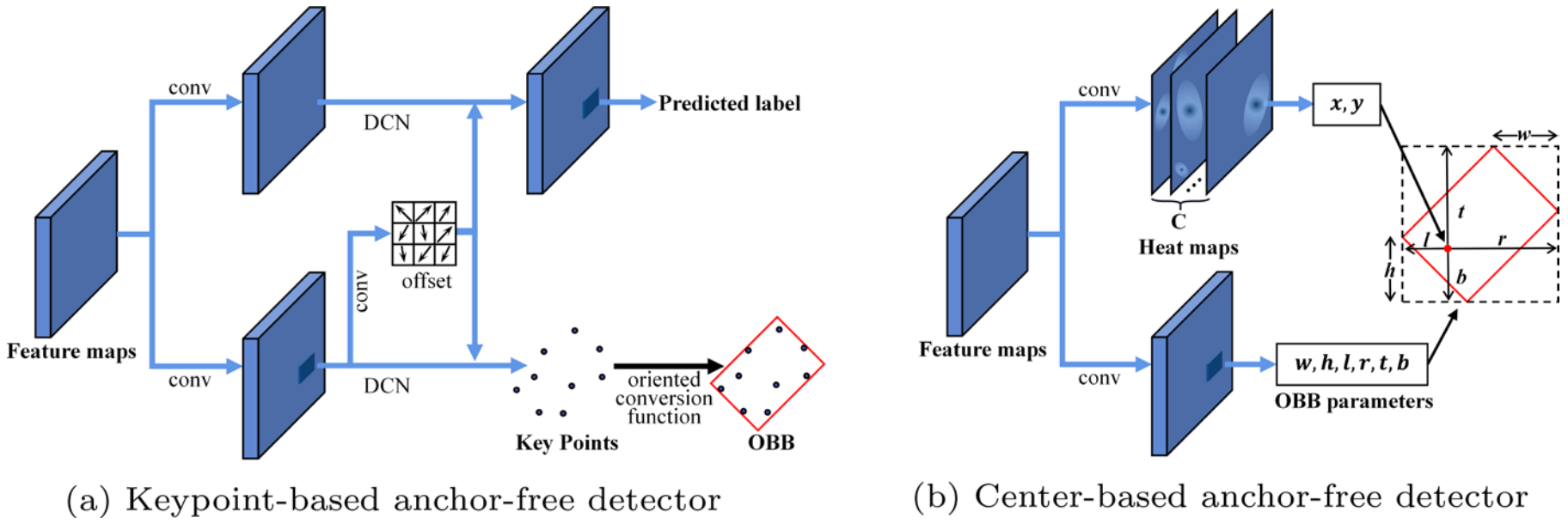

**Fig. 10** The basic architecture of keypoint-based and center-based anchor-free detectors

Center-based methods typically generate multiple probabilistic heatmaps and a series of feature maps. As shown in Fig. 10b, the heatmaps provide a set of candidates (peak points) as coarse center points, whereas the feature maps regress the transformation parameters to represent the OBB. Currently, most center-based methods, including CHPDet (Zhang et al. 2022), GGHL (Huang et al. 2022), and DHRec (Nie and Huang 2023), are dedicated to designing a variety of OBB representations for addressing PoA problems. However, these methods typically follow one-stage paradigms and tend to predict coarse locations due to feature misalignment, whereas the state-of-the-art methods generally involve one or multiple refined stages to improve their performance.

Hence, an effective scheme for increasing the performance of such an approach is to leverage anchor-free methods to generate coarse detection results that are then refined via a subsequent feature alignment stage, e.g., AOPG (Cheng et al. 2022), DEA (Liang et al. 2022), DRDet (Zhang et al. 2023), and TS-Conv (Huang et al. 2024). AOPG (Cheng et al. 2022) initially produces coarse oriented boxes via the rotated FCOS approach (Tian et al. 2019), subsequently refining them into high-quality oriented proposals. DEA (Liang et al. 2022) leverages two parallel branches, which separately generate proposals using anchor-free and anchor-based approaches, followed by an interactive sample screening procedure to select high-quality training samples.

In contrast with the above two methods, which capitalize on the merits of anchor-free techniques to mitigate spatial misalignment and facilitate an appropriate sample assignment process, DRDet (Zhang et al. 2023) and TS-Conv (Huang et al. 2024) concentrate on feature refinement. DRDet (Zhang et al. 2023) adopts two perpendicular rotated lines to represent the OBB. Then, an orientation-guided feature encoder (OFD) is employed to encode the orientation-aware information into refined features along each rotated line. Compared with rectangular features, the line features extracted from the OFD can introduce less noise and alleviate the feature aliasing issue caused by overlapping objects. TS-Conv (Huang et al. 2024) utilizes different sampling offsets for localization and classification to alleviate the task misalignment problem (i.e., localization and classification tasks may focus on different feature regions (Song et al. 2020)). The sampling offsets are restricted by the initial OBBs predicted in an anchor-free manner, allowing for dynamic adaptations to objects with various shapes.

### 3.4 DETR-based detectors

In addition to the above convolution-based methods, DETR-based detectors, including DETR (Carion et al. 2020) and its variants (Zhu et al. 2021; Sun et al. 2021; Gao et al. 2021), have exhibited great potential and achieved state-of-the-art performance in the detection community. On the basis of DETR (Carion et al. 2020), $O^2$DETR (Ma et al. 2021) was proposed to utilize a transformer for completing oriented object detection tasks. In addition, depthwise separable convolutions (Sifre and Mallat 2013; Chollet 2017; Haase and Amthor 2020) were introduced to replace the computationally complex self-attention mechanism, making networks more lightweight and speeding up their training processes. To address feature misalignment, Dai et al. (2022) proposed AO2-DETR by improving the deformable DETR (Zhu et al. 2021); they designed an oriented proposal generation mechanism and an adaptive oriented proposal refinement (OPR) module for aligning the features. Recently, several improved DETR-based detectors have been proposed for generic object detection tasks, e.g., DN-DETR (Li et al. 2022), DAB-DETR (Liu et al. 2022), and DINO (Zhang et al. 2022), resulting in dramatic breakthroughs in terms of accuracy and convergence speed. ARS-DETR (Zeng et al. 2024) attempts to exploit DINO (Zhang et al. 2022) in oriented object detection tasks. Compared with other advanced oriented object detectors, it achieves greater detection accuracy in terms of a more rigorous metric (i.e., $AP_{75}$), but it lags w.r.t. the standard metric (i.e., $AP_{50}$). Moreover, the long training convergence time and heavy computational cost of this method are still open problems.

### 3.5 Discussion

Feature misalignment and spatial misalignment severely impair the performance of oriented object detection methods. To address these issues, substantial research has contributed to modifying the current detection frameworks. The existing two-stage detectors typically employ efficient and precisely oriented proposal generation modules, leveraging RRoI operators to extract rotated aligned features. Similarly, one-stage detectors are inclined to incorporate an extra refined stage for feature alignment. Thus, the above two schemes can empower detectors to mine rotation-related information, thereby enhancing the semantic representations of oriented objects.

Nevertheless, spatial misalignment remains a persistent issue. Anchor-free detectors address this problem by eliminating the anchor mechanism and, in advanced methods, incorporating extra feature refinement stages to further mitigate feature misalignment. In addition, DETR-based methods provide a new detection paradigm and have received widespread attention. However, the exploration of DETR-based methods in oriented object detection is not sufficiently comprehensive, and further research is needed to accelerate the convergence of the training process and reduce the computational overhead.

Well-designed detection frameworks are conducive to alleviating feature and spatial misalignment but fail to address PoA problem. In addition, the extracted features are not equipped with rotation invariance, as the convolution operators are axis-aligned. To overcome these dilemmas, suitable OBB regression techniques and powerful feature representations of oriented objects have also been widely studied since they can be seamlessly integrated into various detection frameworks. Next, we discuss the OBB regression techniques and feature representation approaches.

## 4 OBB regression technologies

Oriented object detectors typically locate objects through regression. Specifically, a customized regression head predicts the orientation parameters when the most frequent $\theta$-based representation is used. Unfortunately, such a regression paradigm suffers from several limitations, including metric-loss inconsistency and angular boundary discontinuities. For more details about the above issues, please refer to Sect. A of the Appendix or the corresponding papers (Qian et al. 2021, 2022; Yang et al. 2021, 2022; Xu et al. 2021). To address these issues, the existing oriented object detection methods usually develop novel loss functions or alternative representations for OBBs. Several representative methods are shown in Fig. 11. Next, we briefly introduce them and discuss their advantages and disadvantages.

### 4.1 Regression loss

(1) *Metric-loss inconsistency* Metric-loss inconsistency generally imply that the optimal choice for the regression task may not guarantee high localization accuracy in terms of the IoU. To bridge this gap, the existing generic object detectors generally introduce IoU-induced loss functions, such as the GIoU (Rezatofighi et al. 2019) and DIoU (Zheng et al. 2020). However, these IoU-induced losses cannot be incorporated directly into oriented object detection because of the nondifferentiable nature of the RIoU (Yang et al. 2021). Thus, several differentiable functions have been designed to approximate the RIoU loss (Yang et al. 2019, 2021, 2022). The PIoU (Chen et al. 2020) introduces a differentiable kernel function that accumulates the contributions of interior overlapping pixels to approximate their intersection area. Several solutions (Yang et al. 2019, 2022, 2021) integrate the RIoU as a loss weight of the regression loss:

$$L_{RIoU} = \frac{L_{reg}}{|L_{reg}|} \cdot |g(RIoU)| \tag{1}$$

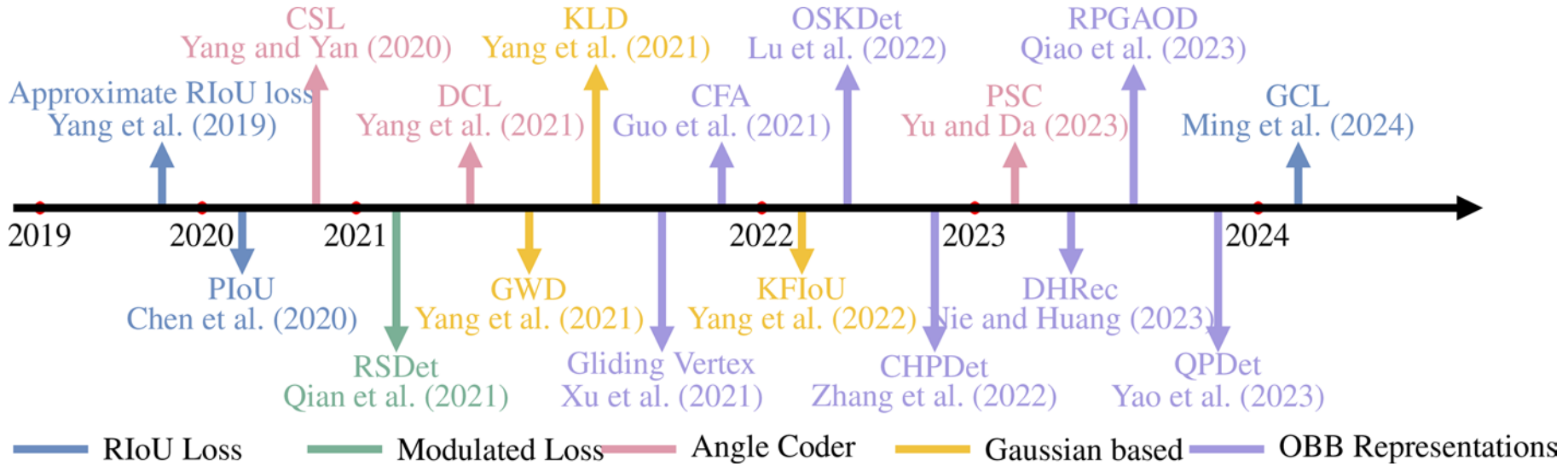

**Fig. 11** Chronological overview of oriented object detection methods for addressing OBB regression problems

$L_{reg}$ denotes the commonly used smooth L1 loss (Ren et al. 2015, 2017). $g(\cdot)$ is a loss function related to the RIoU, e.g., $-log(\cdot)$. This loss is composed of a normalized regression loss $\frac{L_{reg}}{|L_{reg}|}$ for controlling the direction of gradient propagation and a scalar $g(RIoU)$ for adjusting the gradient magnitude. When the RIoU is close to 1, $g(RIoU) \approx 0$, and $L_{reg}$ is approximately equal to 0, effectively mitigating metric-loss inconsistency.

In addition to designing a regression loss for approaching the RIoU, Ming et al. (2024) analyzed the gradient of the RIoU loss and proposed a gradient calibration loss (GCL). The GCL constructs a corrected gradient w.r.t. the RIoU, angular error, and scale and then calculates the optimized regression loss through integration. Despite these efforts, angle regression still faces challenges, particularly the PoA problem.

(2) *Angular boundary discontinuities* Owing to the PoA problem, the regression loss sharply increases when the angle approaches its boundary or the aspect ratio approaches 1, which seriously confuses the utilized network and causes training instability. Thus, several methods have been proposed to address these issues, and they can be divided into three types.

*Modulated rotated loss* (Qian et al. 2021, 2022). The modulated rotated loss adds an extra loss item based on the naïve regression loss to eliminate angular boundary discontinuities. Specifically, it first transforms the original predicted OBB $b_p = (x_p, y_p, w_p, h_p, \theta_p)$ to another form $b_p^{'} = (x_p, y_p, h_p, w_p, \theta_p - \frac{\pi}{2})$,[1] and then it takes the minimum of their regression loss, i.e., $\min\left\{L_{reg}(b_p, b_g), L_{reg}(b_p^{'}, b_g)\right\}$, where $b_g$ denotes the corresponding GT. This scheme can adaptively choose the appropriate representation for the predicted OBB that results in the smallest loss value, thereby mitigating the sudden loss increases near angular boundaries. However, this approach does not fully resolve the metric-loss inconsistency.

*Angle coder* (Yang and Yan 2020; Yang et al. 2021; Yu and Da 2023, 2024). The circular smooth label (CSL) approach discretizes angles into intervals and predicts a discrete angle via classification (Yang and Yan 2020). In addition, to increase its tolerance to error concerning the adjacent angles and to handle the PoA problem, CSL uses a window function for angle label smoothing. Although CSL eliminates boundary discontinuities, its heavy prediction layer harms its efficiency. To address these issues, Yang et al. (2021) further adopted densely coded labels (DCLs) to reduce the code length. Furthermore, Wang et al. (2022) analyzed the limitations of CSL when continuous focal loss functions (Lin et al. 2017, 2020) were directly applied to the soft labels of an angle classification task. Specifically, when the label $y \neq 1$, the extreme point of the derivative of the focal loss function $FL(x)$ is not at $x = y$, yielding an inaccurate angle prediction. Thus, Gaussian focal-CSL (GF-CSL) was designed to obtain more accurate angle predictions with higher responses at the peaks by implementing adaptive Gaussian attenuation on the negative angle categories. However, the hyperparameters of these methods have significant effects on their performance. Even worse, the optimal settings for different datasets also differ, thereby requiring laborious tuning processes.

To solve this problem, Yu and Da (2023, 2024) designed a differentiable angle coder named the phase-shifting encoder (PSC). The PSC encodes the angle into a periodic phase to solve the boundary discontinuity problem. Moreover, an advanced version, PSCD, maps angles to phases with different frequencies to further solve the square-like problem.

[1] Notably, the modulated rotated loss is customized for representing OBBs under the OpenCV definition, which is an intractable problem to the exchangeability of edges and the PoA.

*Gaussian distribution-based methods* (Yang et al. 2021, 2022, 2021, 2022). Gaussian distribution-based methods provide unified and elegant solutions to the boundary discontinuities and the square-like problem. First, the OBB representation $b = (x, y, w, h, \theta)$ is converted to a 2-D Gaussian distribution representation $\mathcal{N}(m, \Sigma)$, where $m = (x, y)$ and $\Sigma$ is a matrix associated with $w, h, \theta$:

$$\Sigma^{\frac{1}{2}} = \begin{bmatrix} \frac{w}{2}\cos^2\theta + \frac{h}{2}\sin^2\theta & \frac{w-h}{2}\cos\theta\sin\theta \\ \frac{w-h}{2}\cos\theta\sin\theta & \frac{h}{2}\cos^2\theta + \frac{w}{2}\sin^2\theta \end{bmatrix} \tag{2}$$

A distance function is subsequently used to measure two Gaussian distributions, such as the Gaussian Wasserstein distance (GWD) (Yang et al. 2021) or the Kullback–Leibler divergence (KLD) (Yang et al. 2021). Furthermore, the measure is converted into an approximate IoU loss by using a nonlinear transformation to achieve metric-loss consistency. The merit of the Gaussian distribution is that the angle is encoded by a trigonometric function and is thereby not constrained by the PoA problem. Moreover, the OBB parameters are jointly optimized in a dynamic manner so that they can influence each other during training. Despite their advantages, both the GWD and KLD maintain only value-level rather than trend-level metric-loss consistency. To achieve better trend-level alignment, the KFIoU loss was proposed; this approach is differentiable and does not require additional hyperparameters. The KFIoU can calculate the overlapping area between two Gaussian distributions, resulting in improved performance relative to that of the GWD and KLD.

## 4.2 OBB representations

The most straightforward approach for handling angular boundary discontinuities is to design a novel OBB representation. One common scheme exploits new parameters to redefine the OBB, e.g., quadrilateral representation (Xu et al. 2021) and DHRec (Nie and Huang 2023). The quadrilateral representation adopts the coordinates of four vertices to represent an OBB but suffers from inconsistency when conducting vertex sorting[2] between the predictions and the GT, as shown in Fig. 12a. RSDet (Qian et al. 2021, 2022) mitigates this problem by introducing a modulated loss that considers different vertex orderings, minimizing the regression loss across these variations. Furthermore, Xu et al. (2021) proposed an effective way to glide the vertex of a horizontal anchor on each corresponding side, as shown in Fig. 12b. Specifically, it regresses four length ratios representing the gliding offset on each corresponding side, eliminating the confusion caused by vertex sorting. To remedy the confusion encountered in cases with nearly horizontal objects, it directly selects horizontal detection guided by the predicted obliquity factor, albeit with slightly imprecise regression results. DHRec (Nie and Huang 2023) encodes the OBB using double HBBs derived from the sorted horizontal and vertical coordinates, as shown in Fig. 12c. Hence, such a method allows any horizontal object detector to be harnessed to predict OBBs.

In addition, several anchor-free methods utilize keypoints to denote an OBB, which can provide rich semantic features for oriented objects. Following RepPoints (Yang et al. 2019), CFA (Guo et al. 2021, 2022) and Oriented RepPoints (Li et al. 2022) utilize deformable convolution (Dai et al. 2017) to generate a group of reppoints, as shown in Fig. 12d. A minimum bounding rectangle is then computed for each set of predicted reppoints to yield detection results. OSKDet (Lu et al. 2022) encodes 8 ordered points (4 vertices and 4 midpoints) to represent an OBB because the object has more obvious features in its vertex and edge areas, as shown in Fig. 12e. Furthermore, an orientation-sensitive heatmap was designed to better fit object shapes, allowing the utilized model to implicitly learn orientations and shapes.

[2] For more details, please refer to Sect. B of the Appendix.

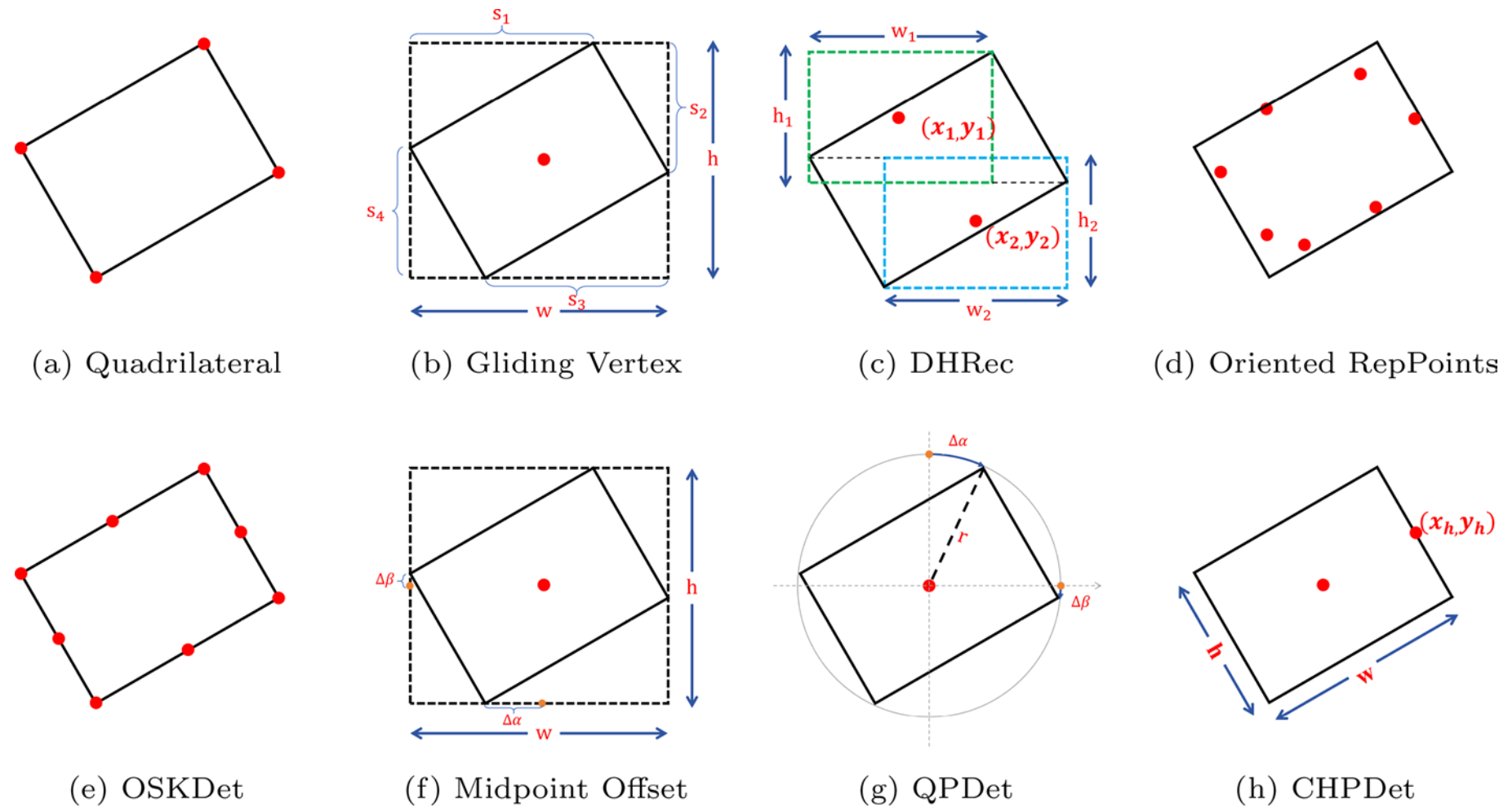

**Fig. 12** Comparison of different OBB representation methods. The red dots and parameters denote corresponding OBB parameters

Although angular discontinuities have been eliminated, the redesigned OBB representation still have limitations. The quadrilateral representation is irregular, and DHRec adds extra position and oblique factors to ensure the uniqueness of the produced OBB representation. Moreover, the keypoints rely on a complicated post-processing operator to generate a rectangular box.

Therefore, another approach for OBB representation only uses some extra parameters to determine the orientations of objects. The classic midpoint offset representation (Xie et al. 2021) infers orientations via midpoint offsets, as shown in Fig. 12f, but it typically produces parallelograms that require regularization. On the basis of the midpoint offset representation, Qiao et al. (2023) analyzed the geometric relationship between an OBB and its external HBB to derive the height directly from the width and two offsets, using only five parameters to generate a high-quality OBB directly from the horizontal anchors. QPDet (Yao et al. 2023) adopts two symmetrical offsets w.r.t. the quadrant points to account for the rotation and aspect ratio, and a single parameter (radius $r$) controls the scale, as shown in Fig. 12g. CHPDet (Zhang et al. 2022) defines a head point to indicate the correct orientation but requires proper annotations that specify the direction of the object head within the range of $2\pi$, as shown in Fig. 12h. These orientation representation can discard angle regression, thus naturally eliminating angular boundary discontinuities. Additionally, the advantages and disadvantages of the above OBB representation methods are summarized in Table 2.

### 4.3 Discussion

As stated above, an enormous amount of research effort is committed to resolving the challenges encountered by OBB regression. Redesigning a novel regression loss for mainstream $\theta$-based representations empowers the detector to solve the metric-loss inconsistency problem and eliminate the confusion caused by the PoA, thereby enhancing the stability of the network backpropagation process. In particular, Gaussian distribution-based methods draw upon a trigonometric encoder and joint optimization to achieve strong performance. On the other hand, novel OBB representation approaches can avoid performing orientation regression, in which completely redefined OBB representations commonly rely on complex post-processing operations or extra constraints, whereas the orientation representations provide a simple yet efficient way to determine the orientation. Nevertheless, only a handful of novel OBB representation approaches consider the inconsistency problem.

**Table 2** Comparison among different OBB representation methods

| Methods | Advantages | Disadvantages |
|---|---|---|
| Quadrilateral representation (Xia et al. 2018; Qian et al. 2021, 2022) | This is a direct method that can compactly enclose oriented objects with large degrees of deformation and has been widely adopted to annotate objects in large-scale RS datasets | It suffers from an inconsistent vertex sorting process, and can only represent irregular quadrilaterals (not rectangles) |
| Gliding Vertex (Xu et al. 2021) | This is a concise but effective method that can eliminate vertex sorting | It can only represent irregular quadrilaterals (not rectangles). It cannot accurately represent nearly horizontal objects |
| DHRec (Nie and Huang 2023) | It can directly use horizontal object detectors to regress OBBs, thus freeing it from the trouble related to the PoA and vertex sorting problems | It requires too many parameters to ensure the uniqueness of OBB representations |
| CFA and Oriented RepPoints (Guo et al. 2021, 2022; Li et al. 2022) | The adaptive point learning method can capture the geometric information of oriented objects and avoid vertex sorting and the PoA problems | It requires complicated post-processing operations to convert reppoints into an OBB |
| OSKDet (Lu et al. 2022) | The keypoints can capture the critical features of vertex and edge areas, which can better match the object shape. The unordered keypoint representation scheme can avoid the confusion of vertex sorting | It requires complicated post-processing operations to convert irregular keypoints into an OBB |
| Midpoint Offset (Xie et al. 2021) | This is a concise yet effective OBB representation method that can eliminate the PoA problem | It typically generates a parallelogram and requires a post-processing procedure for regularization |
| QPDet (Yao et al. 2023) | This is a simple OBB representation method that needs just five parameters; it can avoid the generation of irregular bounding boxes | It suffers from a new PoA problem between $\Delta\alpha$ and $\Delta\beta$ when representing nearly horizontal objects |
| CHPDet (Zhang et al. 2022) | This approach employs a head point to indicate the direction, effectively eliminating the PoA problem | The accuracy of the orientation depends on the precision of the center point and the head point. When either position is inaccurate, it affects the overall precision of the predicted location |

## 5 Feature representation approaches

Robust and discriminative feature representations play pivotal roles in improving both localization and classification tasks. As a result, the most recent improvements in detection accuracy have been attained via research conducted on enhancing feature representations through innovative network architectures. In this section, we review the effort devoted to improving the feature representations of oriented objects, i.e., rotation-invariant feature representations and advanced feature representations. Several key methods are shown in Fig. 13.

### 5.1 Rotation-invariant feature representations

Rotation invariance is an essential problem when learning visual feature representations for oriented objects (Lowe 2004; Han et al. 2021; Yu et al. 2024). The commonly used approaches, including RRoI operators (Ma et al. 2018; Yang et al. 2018; Ding et al. 2019) and random rotation data augmentation (Han et al. 2021), are suboptimal, as they can extract only approximately rotation-invariant features (Lenc and Vedaldi 2015; Worrall et al. 2017).

Recently, the exploration of rotation-sensitive feature extraction networks has provided new insights for the community; these methods utilize different channels to represent feature information derived from different orientations, e.g., oriented response convolution (ORConv) (Zhou et al. 2017) and group-equivariant convolutional neural networks (G-CNNs) (Cohen and Welling 2016; Worrall et al. 2017; Marcos et al. 2017; Weiler and Cesa 2019; Weiler et al. 2018). Several methods, e.g., RRD (Liao et al. 2018), CHP-Det (Zhang et al. 2022), and S$^2$ANet (Han et al. 2022), replace ordinary convolution modules with ORConv to obtain orientation-dependent responses, which are then transformed into rotation-invariant features using ORAlign and ORPooling. Additionally, ReDet (Han et al. 2021) incorporates G-CNN into RoI Transformer (Ding et al. 2019) to generate rotation-equivariant features; then, a rotation-invariant RoI alignment operator is designed to adaptively extract rotation-invariant features from the equivariant features according to the predicted orientations. Furthermore, Li et al. (2024) introduced selective rotation of the kernel (SRK) to enhance classification features. The SRK module rotates the convolution kernel at different angles to extract rotation-invariant features, where the output channel dimensionality of the corresponding rotated convolution kernel is adaptively obtained by network learning.

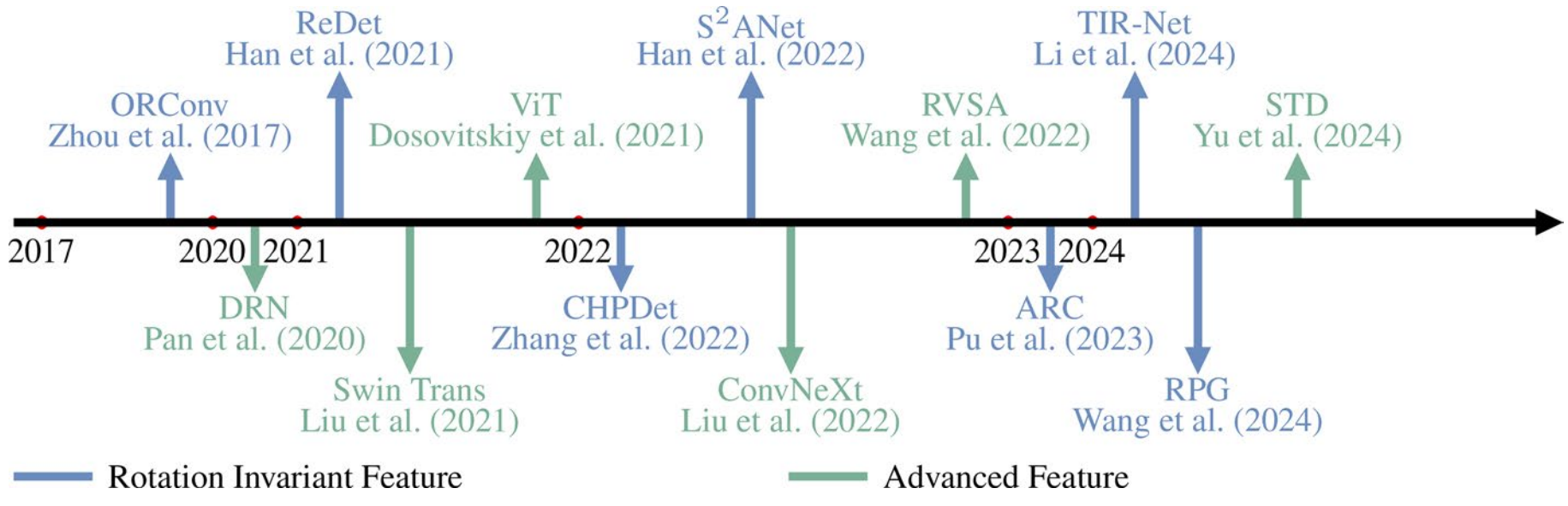

**Fig. 13** Chronological overview of feature representation approaches

In contrast with ORConv and G-CNNs, which extract features via static kernels with a group of fixed orientations, ARC (Pu et al. 2023) dynamically rotates its convolution kernels according to the orientations of the objects, where these orientations are predicted in a data-dependent manner. As a result, ARC can capture the rotation information of objects with different orientations and boost the feature representations produced in oriented object detection scenarios.

However, Wang et al. (2024) noted that the cyclic shift phenomena exhibited by the features acquired from a rotation-equivariant network are unstable due to dilation and pooling operations. To maintain rotation invariance for features, a rotation-robust prototype generation (RPG) scheme was designed with a stabilization module and an enhancement module. The former aggregates features with different orientations to generate a rotation-robust prototype, whereas the latter employs this prototype to enhance the original features in each group.

## 5.2 Advanced feature representations

In addition to rotation-invariant networks, advanced feature extraction networks, particularly vision transformers (ViTs) (Vaswani et al. 2017; Dosovitskiy et al. 2021; Liu et al. 2021), have played a crucial role in achieving high-precision detection. ConvNeXt (Liu et al. 2022) is another notable architecture that has also contributed to the field, although the corresponding discussion will be brief in this context.

Recently, ViTs have achieved significant success in computer vision (Dosovitskiy et al. 2021; Liu et al. 2021; Han et al. 2023), primarily because their self-attention mechanisms capture global feature representations. This exceptional feature representation capacity has led to the increasing adoption of ViTs in object detection tasks, yielding remarkable results. Representative architectures such as the ViT series (Dosovitskiy et al. 2021; Xu et al. 2021; Zhang et al. 2023) and the Swin transformer (Liu et al. 2021) can serve directly as backbone networks, exhibiting better feature representation capabilities than CNNs do. Furthermore, the unsupervised MAE-based (He et al. 2022) pretraining scheme has made notable progress in terms of developing ViTs for object detection. Consequently, these powerful ViT architectures contribute to establishing a solid foundation for delivering outstanding achievements in oriented object detection tasks.

Nevertheless, their utilization in oriented object detection field is fairly unexplored; e.g., an essential problem is how to extract rotation-related features. Wang et al. (2022) addressed this concern by designing a rotated variable-size window attention (RVSA) mechanism based on a ViT, which adaptively generates locally oriented windows with different sizes, locations, and angles. Although RVSA outperforms all the previously proposed methods, it relies on a self-attention mechanism to create oriented windows without explicitly leveraging guidance information.

On the other hand, STD (Yu et al. 2024) adopts a controlled scheme for manipulating the feature extraction process according to the decoupled OBB parameters, i.e., the center position, sizes, and angles. It follows a divide-and-conquer approach that estimates the position, size, and angle parameters via separate network branches in different stages. Furthermore, the cascaded activation masks created by the decoupled OBB parameters are integrated to gradually enhance the features extracted by stacked transformer blocks. The progressive refinement of feature representations enables STD to reach state-of-the-art performance, achieving mean average precision (mAP) values of 82.24% and 98.55% on the DOTA-V1.0 (Xia et al. 2018) and HRSC2016 (Liu et al. 2016) datasets, respectively.

More recently, ConvNeXt (Liu et al. 2022) gradually modified the standard ResNet model according to a series of design decisions made by the Swin transformer (Liu et al. 2021) and demonstrated that pure CNNs outperform ViTs in terms of accuracy and robustness. Moreover, ConvNeXt can maintain the efficiency of standard CNNs, thus becoming the dominant architecture in many applications.

### 5.3 Discussion

The investigation of feature representations can lead to improvements in the whole object detection field. ORConv and G-CNNs empower the corresponding models to mine rotation-invariant features by using different channels to represent feature information derived from different orientations, whereas advanced feature extraction networks are dedicated to enhancing semantic representations via powerful and well-designed architectures. Although the former is conducive to extracting rotation-invariant features in both the spatial and channel dimensions, they are built on conventional CNN modules that lag behind the latter. Thus, it is crucial to validate the effectiveness of integrating rotation-invariant feature extraction networks and advanced networks. We hope that further research efforts will explore more powerful rotation-invariant and high-level semantic feature representations for performing oriented object detection.

## 6 Solutions to common issues

In addition to the specific challenges associated with oriented object detection, several common issues still exist w.r.t. RS scenarios, e.g., complex backgrounds, scale variations, large aspect ratios, and the lack of annotated samples, as shown in Fig. 14. Several representative methods for tracking these common issues are shown in Fig. 15.

### 6.1 Complex backgrounds

Owing to their wide visual fields and complex earth surfaces, RS images typically contain a variety of complex backgrounds, causing significant interference in detection tasks. Objects are frequently surrounded by different backgrounds, necessitating detectors with heightened discriminative capabilities. In addition, the presence of backgrounds with textures and shapes resembling the objects leads to a high false-positive rate.

To this end, a series of efforts have been made to suppress background noise and emphasize the valuable areas of objects. CFC-Net (Ming et al. 2022) and PETDet (Li et al. 2024) both combine channel and spatial attention modules to learn the semantic correlation between the foreground and background of the image. However, these methods rely on self-attention mechanisms and lack direct foreground guidance. To enhance the ability to discriminate the foreground, Zhang et al. (2022) proposed a foreground relation module for obtaining foreground contextual representations under the supervision of the designed foreground map. Similarly, CBDA-Net (Liu et al. 2022) builds two parallel spatial attention streams to capture center and boundary attention features, which can assist detectors in improving their object localization accuracy. In addition, SCRDet (Yang et al. 2019) adopts a pixel attention network to generate a saliency map that can separate the foreground from the background and uses squeeze-and-excitation (SE) blocks (Hu et al. 2020) as its channel attention network to further enhance the saliency map.

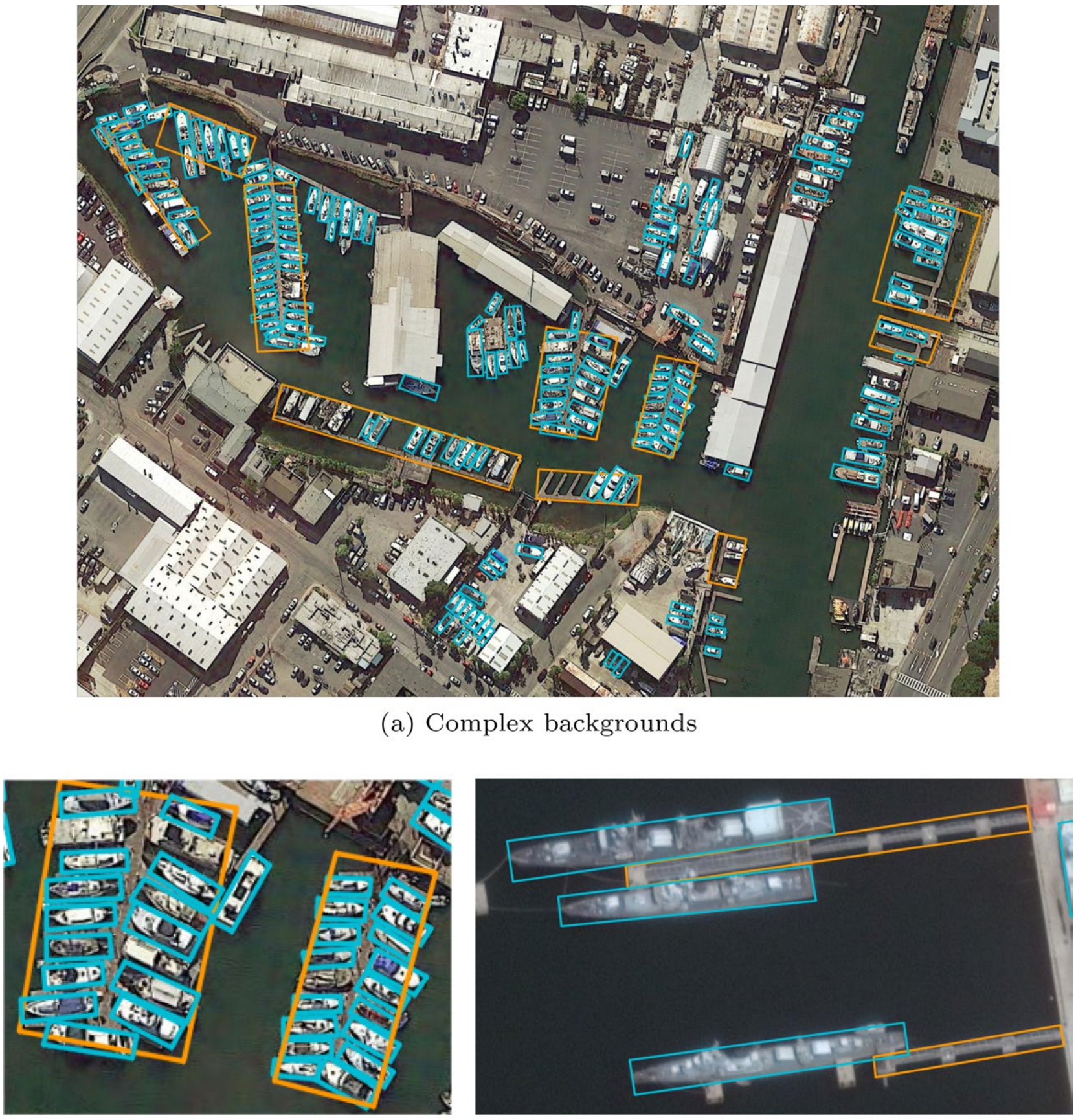

(a) Complex backgrounds

(b) Scale variations

(c) Large aspect ratio

**Fig. 14** Illustration of common issues in RS scenarios (Xia et al. 2018; Ding et al. 2022). **a**–**c** Examples of complex backgrounds, scale variations, and large aspect ratio, respectively

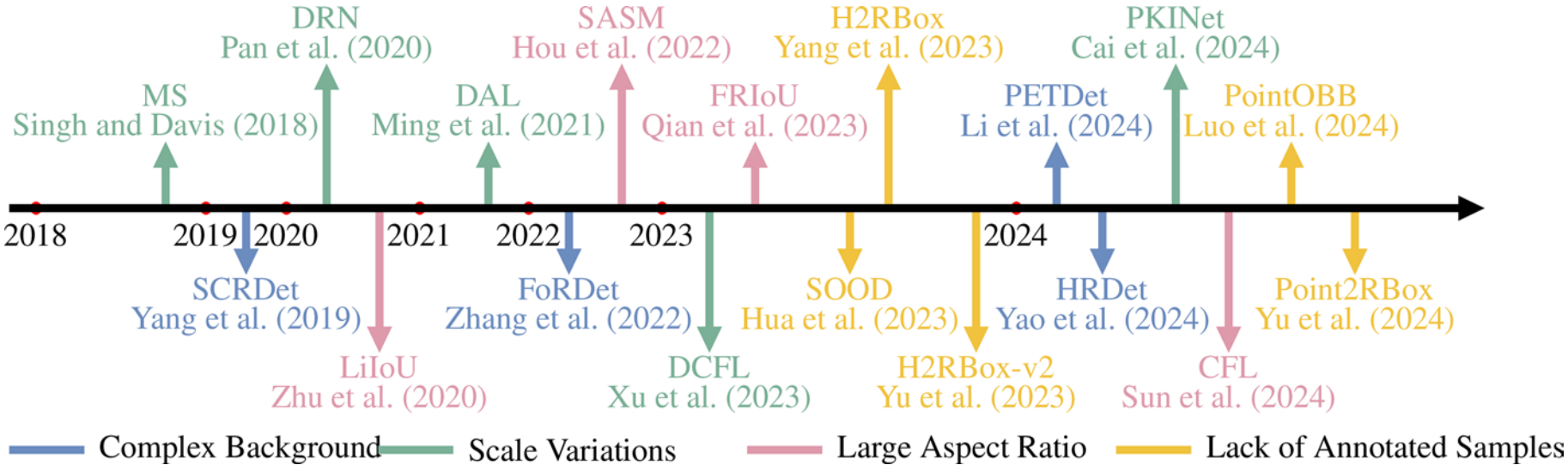

**Fig. 15** Chronological overview of oriented object detection methods for addressing common issues

Building on SCRDet, SCRDet++ (Yang et al. 2022) introduces an instance-level denoising (InLD) module that weakens the feature response of the background region while decoupling the features of different categories into their corresponding channels. Similar to InLD, several methods (Zhang et al. 2023; Yu et al. 2023; Yao et al. 2024; Zheng et al. 2024) introduce semantic mask modules that are supervised by classwise masks transformed from the oriented GT. These semantic mask modules separate the features of different categories in the channel dimension, which can help the networks reduce both background interference and interclass interference.

## 6.2 Scale variations

As the ground sampling distance (GSD) can range from a few centimeters to hundreds of meters, the RS images taken by different sensors usually have large-scale variations. Additionally, even within the same category, object instances exhibit wide-ranging sizes. These interclass and intraclass scale variations pose additional challenges. In particular, the most critical challenge concerns small objects because of their insufficient feature information, inaccurate localization effects, and inadequate positive samples (Cheng et al. 2023). Worse still, the task concerning oriented objects becomes more difficult because of the extra orientation regression process and limited overlaps with anchor boxes.

In recent years, many effective strategies have been developed to increase the robustness and adaptability of detectors to objects with various scales; these approaches can be classified into two categories: network-level methods and data-level methods. Network-level methods develop novel network structures for multiscale feature extraction, e.g., feature pyramid architectures (Lin et al. 2017) and their variants (Tian et al. 2024), as well as multibranch architectures (Li et al. 2019; Pan et al. 2020; Cai et al. 2024). Pan et al. (2020) designed a feature selection module to adjust receptive fields; they proposed a channel attention network to adaptively fuse the features extracted by using kernels with various sizes, aspect ratios, and orientations. PKINet (Cai et al. 2024) employs parallel multiscale convolution kernels without dilation to effectively capture features across various receptive fields. On the other hand, data-level methods strive to design data augmentation strategies that are independent of the applied network architectures and can be generalized to any detector. Multiscale training and testing is a useful data augmentation approach that scales input images at different resolutions (Singh and Davis 2018; Singh et al. 2018) and has been shown to reduce overfitting and improve generalizability (Ren et al. 2015, 2017; Russakovsky et al. 2015). However, this approach inevitably leads to poor temporal efficiency.

In addition to enhancing semantic representations or amplifying object sizes, several optimal assignment strategies have emerged to enable adequate sample assignments for small oriented objects. Anchor-free methods can generate more positive samples of small objects that are apt to be ignored in anchor-based methods. Consequently, approaches such as APE (Zhu et al. 2020) and AOPG (Cheng et al. 2022) generate samples in an anchor-free manner, refining them for obtaining high-quality detection results. Liang et al. (2022) proposed a dynamic enhancement-based anchor network that combines the advantages of anchor-free and anchor-based methods and uses an interactive sample screening procedure to yield higher-quality training samples.

Rather than relying on the IoU alone, several methods (Ming et al. 2021; Zhang et al. 2023; Fu et al. 2024; Huang et al. 2024) use the prior (e.g., the IoU between the anchor and GT) and posterior information (e.g., classification and localization confidences) as the evaluation criteria for screening positive samples. This scheme can push detectors to dynamically mine high-quality anchors as positive samples. In contrast to static priors (e.g., anchor boxes or points) with fixed strides, Xu et al. (2023) designed a dynamic prior with a coarse-to-fine assigner. Specifically, it first uses deformable convolution (Dai et al. 2017) to adaptively adjust the prior location and then leverages coarse prior matching and a finer posterior constraint to dynamically assign samples. This strategy can adaptively assign positive or negative samples according to the shape and posterior information of the objects, increasing the performance achieved by mainstream detectors for small objects.

## 6.3 Large aspect ratios

RS images frequently encompass several categories with extremely large aspect ratios, such as bridges, ships, and harbors. The RIoUs between these categories with anchors are highly sensitive to orientation errors, thereby causing two primary challenges, i.e., spatial misalignment and inaccurate localization.

Accordingly, a series of well-designed assignment strategies and regression losses have been proposed to remedy the corresponding issues. Zhu et al. (2020) introduced a length-independent IoU (LiIoU). The LiIoU intercepts part of the object box along its long side based on the length of the anchor and subsequently calculates the IoU between the intercepted object box and the anchor. Compared with the conventional IoU, the LiIoU facilitates the assignment of more positive samples to long objects, thereby increasing the recall rate. Qian et al. (2023) emphasized that the sampling locations of positive samples should be close to the center distribution of the oriented GT. This is because two horizontal anchors, despite having the same IoU as that of the external HBB of an oriented GT with a large aspect ratio, can differ considerably in their coverage of the GT. Consequently, Qian et al. (2023) and Wang et al. (2024) both combined the horizontal IoU and sample feature alignment overlap degree to evaluate the quality of an anchor, where the sample feature alignment overlap degree is defined as the ratio of the intersection between the horizontal anchor and the oriented GT to the GT itself, reflecting the overlap proportions of oriented object features.

Furthermore, inspired by ATSS (Zhang et al. 2020), SASM (Hou et al. 2022) and CFL (Sun et al. 2024) use a monotonically decreasing function of the aspect ratio as a weight for an IoU threshold that controls the sample assignment process. This approach allows long objects to be assigned lower IoU thresholds. Additionally, several methods (Liu et al. 2022; Qiao et al. 2023; Gong et al. 2024; Xie et al. 2024) construct weighted orientation losses that depend on the aspect ratio, effectively mitigating the effect of the aspect ratio on orientation regression.

### 6.4 Lack of orientation-annotated samples

Owing to its reliance on numerous and laborious annotations, oriented object detection has experienced significant advancements in recent years. However, OBB annotation is a time-consuming and expensive process, resulting in many detection datasets that use HBB annotations but do not provide OBB annotations, thereby restricting their potential application scopes. To alleviate the annotation burden, researchers have explored two avenues: weakly supervised learning, in which OBB annotations are replaced by HBB-level or even point-level annotations; and semisupervised learning, in which only a few images from the whole training dataset are annotated.

The mainstream semisupervised oriented object detection methods commonly follow the pseudo-labeling framework, which consists of a teacher model and a student model. The teacher model, which is an exponential moving average (EMA) (Tarvainen and Valpola 2017) of the student model at different historical training iterations, generates pseudo-labels for unannotated images. The paired models are iteratively trained via the following steps. The teacher model provides pseudo-labels for the unannotated images in a batch, whereas the student model makes predictions for both the annotated and unannotated images. Then, the loss for the predictions of the student model is computed. However, the unsupervised nature of the teacher model introduces noise that can mislead the training process of the student model, especially considering the arbitrary orientations of objects, which further impact the resulting pseudo-label quality. Therefore, the current semisupervised oriented object detection methods (Hua et al. 2023; Wang et al. 2024; Wu et al. 2024) are committed to generating high-quality pseudo-labels.

SOOD (Hua et al. 2023) introduces two loss functions, i.e., a rotation-aware adaptive weighting (RAW) loss and a global consistency (GC) loss. The RAW loss focuses on the orientation consistency between each pseudo-label-prediction pair, dynamically weighting each pair based on their orientation gap. The GC loss measures the global similarity between the pseudo-labels and the predictions, effectively mitigates noise disturbances and implicitly regularizes the object relations. Wang et al. (2024) provided an in-depth analysis of the limitations of conventional pseudo-label- and dense pseudo-label-based methods (Zhou et al. 2022). The former approaches adopt a fixed threshold, whereas the latter methods use a fixed quantity, both of which fail to adaptively select high-quality pseudo-labels. To address this issue, global focal learning was proposed to judge important regions on the basis of the difference between the predictions of the teacher model and the student model, guiding networks to focus more on inconsistent regions during training. In addition, the Pseudo-Siamese Teacher (Wu et al. 2024) adopts two teacher models that are updated by different optimization schemes to improve the reliability of pseudo-labels and uses the Jensen-Shannon divergence measure to eliminate inconsistent pseudo-labels.

The mainstream weakly supervised oriented object detection methods commonly consist of multiple branches that are fed with multiple augmented views of the input image. Various consistent losses are then designed to align the features or predictions derived from different views. H2RBox (Yang et al. 2023), which was the first HBB annotation-based weakly supervised method, follows a weakly and self-supervised angle learning paradigm. The weakly supervised part calculates the regression loss between the external HBB of the predicted OBB and the horizontal GT, whereas the self-supervised part measures the consistency of the angles predicted from two views with different rotation augmentations. On the basis of H2RBox, H2RBox-v2 (Yu et al. 2023) leverages reflection symmetry to learn the orientations of objects in a self-supervised manner. Furthermore, a CircumIoU loss is designed, allowing H2RBox-v2 to be compatible with the random rotation augmentation method.

Compared with OBB and HBB annotations, point annotations have lower costs and greater efficiency.[3] The main challenge faced by the point annotation-based methods lies in enabling the models to perceive the orientations and scales of objects on the basis of point annotations. PointOBB (Luo et al. 2024) was designed with a resized view (by random scaling) and a rot/flp view (implemented by random rotation or vertical flipping) based on original view. Upon these three views, a scale augmentation module and an angle acquisition module were constructed. The former aims to perceive object scales by improving the consistency between the predicted scores of the original and resized views, whereas the latter incorporates self-supervised angle learning to predict angles.

Furthermore, Yu et al. (2024) presented Point2RBox, including synthetic pattern knowledge combination and self-supervised transformation schemes. The former first generates synthetic patterns with known boxes by sampling around each labeled point and then overlays these patterns on the original image, providing knowledge that enables the network to estimate sizes and angles. The latter is similar to PointOBB, which uses original and transformed (randomly selected from rotated, flipped, and scaled) views to perceive the size and orientation differences between objects. In addition, Zhang et al. (2024) proposed a progressive method, named point-to-mask-to-HBB-to-OBB (PMHO), to achieve oriented object detection. However, this framework is time-consuming, with each component being optimized independently; thus, it relies heavily on the capabilities of well-trained models such as the SAM (Kirillov et al. 2023).

## 6.5 Discussion

Complex backgrounds, scale variations, large aspect ratios, and the lack of oriented-annotated samples are crucial issues encountered in RS object detection tasks, and they become more severe in oriented object detection tasks. As stated above, numerous methods have been proposed to address these issues from various perspectives, e.g., data augmentation schemes, assignment strategies, reweighted orientation losses, attention mechanisms, self-supervised losses, and pseudo-labeling frameworks. Unfortunately, the exploration of solutions for these issues is far from mature, so further research may be beneficial. For example, a significant performance gap remains w.r.t. detecting small or long objects compared to normal objects, even for the state-of-the-art detectors. On the other hand, the general split-and-detect scheme is inefficient during inference because too many empty patches that contain only background information are used. Several prior works have preliminarily considered these points, e.g., superresolution-based object detection (Shermeyer and Van Etten 2019; Liu et al. 2023; Zhang et al. 2023) and focus-and-detect schemes (Duan et al. 2021; Koyun et al. 2022). Additionally, the accuracy of semi-/weakly supervised oriented object detection is still far from satisfactory, lagging significantly behind that of fully supervised methods. The combination of weakly supervised and semisupervised methods may lead to new breakthroughs (Wu et al. 2024).

[3] According to https://cloud.google.com/ai-platform/data-labeling/pricing, the cost of point annotations is approximately 50.0% lower than that of HBB annotations and 104.8% lower than that of OBB annotations, and their time consumption is only 1.2x greater than that of image-level annotations.

# 7 Evaluation protocols and datasets

## 7.1 Evaluation protocols

Accuracy and efficiency are both crucial criteria for evaluating the performance of oriented object detectors. The evaluation protocol for OBBs is slightly different from that used for HBBs, as the IoU is replaced with the RIoU. Efficiency evaluations use the frames-per-second (FPS) metric, which is defined as the number of image frames processed by a detector per second, whereas accuracy evaluations account for both precision and recall. The most universally agreed-upon metric for accuracy evaluations is AP.

For the object detection task, the detector outputs $M$ predicted results $\{(b_j, c_j, s_j)\}_{j=1}^{M}$, where each item contains an OBB $b_j$ and a category label $c_j$ with a corresponding confidence score $s_j$. Then, the predicted results are assigned to the GT objects $\{(b_k^*, c_k^*)\}_{k=1}^{N}$ based on the RIoU and category, where $b_k^*$, $c_k^*$ and the superscript $*$ denote the OBB, category label, and GT, respectively. A predicted result $(b_j, c_j, s_j)$, which is assigned a GT object $(b_k^*, c_k^*)$, is judged to be a true positive (TP) if the following criteria are met.

(1) The predicted label $c_j$ accords with the label $c_k^*$ of the GT object.

(2) The RIoU between the predicted OBB $b_j$ and the GT OBB $b_k^*$, denoted by the RIoU $(b, b^*)$, is not smaller than the predefined RIoU threshold $T_{RIoU}$. Otherwise, it is regarded as a false positive (FP).

Once the numbers of TPs and FPs have been obtained, the precision and recall metrics can be calculated. Precision is the proportion of correctly predicted instances among the total number of predicted results, whereas recall is the proportion of all positive instances predicted by the detector among the total number of GT objects. Their formulas are defined as follows:

$$Prec(T_s) = \frac{N_{TP}}{N_{TP} + N_{FP}} \tag{3}$$

$$Rec(T_s) = \frac{N_{TP}}{N_{TP} + N_{FN}} = \frac{N_{TP}}{N} \tag{4}$$

where $N_{TP}$, $N_{FP}$, and $N_{FN}$ denote the numbers of TPs, FPs, and false negatives (FNs), respectively, which are determined by the score thresholds $T_s$ and $T_{RIoU}$. Note that the precision and the recall metrics are functions of the confidence threshold $T_s$ with a fixed $T_{RIoU}$.

However, neither precision nor recall can independently evaluate the accuracy of a detector, whereas AP can combine both precision and recall. For each category, by gradually varying $T_s$ from 1.0 to 0.0, the recall increases as $N_{TP}$ increases, and a list of pairs (*Prec*, *Rec*) can be obtained. This allows precision to be considered a discrete function of recall, i.e., the precision–recall curve (PRC), which is denoted by *P*(*R*). The AP value is obtained by computing the average precision value *P*(*R*) over the interval from $R = 0.0$ to $R = 1.0$:

$$AP = \frac{1}{N} \sum_{n=0}^{Rec(0)} \max_{R \geq \frac{n}{N}} P(R) \tag{5}$$

Ultimately, to evaluate the overall accuracy across all categories, the mAP averaged over all categories is adopted as the final evaluation metric.

### 7.2 Datasets

Recently, several research groups have released dozens of high-quality RS image datasets, each of which dramatically boosts the development of RS object detection methods. Datasets annotated only with HBBs, including DIOR (Li et al. 2020), LEVIR (Zou and Shi 2018), NWPU VHR-10 (Cheng et al. 2014), RSOD (Xiao et al. 2015; Long et al. 2017), xView (Lam et al. 2018), and HRRSD (Zhang et al. 2019), are not covered here. In addition, several oriented object detection datasets with horizontal views, e.g., text detection datasets (Karatzas et al. 2015), and datasets derived from different modalities, such as SAR datasets (Wei et al. 2020; Lei et al. 2021), exhibit significant differences in their viewing angles, scenes, and imaging characteristics relative to optical RS datasets. Thus, this paper does not introduce these datasets. In this subsection, we only focus on introducing optical RS datasets annotated with OBBs, including SZTAKI-INRIA (Benedek et al. 2012), 3K vehicle (Liu and Mattyus 2015), UCAS-AOD (Zhu et al. 2015), VEDAI (Razakarivony and Jurie 2016), HRSC2016 (Liu et al. 2016), DOTA (Xia et al. 2018; Ding et al. 2022), ShipRSImageNet (Zhang et al. 2021), DIOR-R (Cheng et al. 2022), DroneVehicle (Sun et al. 2022), FAIR1M (Sun et al. 2022), and GLH-Bridge (Li et al. 2024). Table 3 shows the parameters of the above RS-oriented object detection datasets for making an intuitive comparison. Given that the emergence of DOTA has greatly promoted the development of oriented object detection, we divide the datasets into two parts, namely, early and modern datasets, based on the time at which DOTA was introduced. Only the most typical datasets among the above datasets are described in detail owing to space restrictions. For more details, please refer to Sect. C of the Appendix.

*DOTA* (Xia et al. 2018; Ding et al. 2022) contains large quantities of objects with a considerable variety of orientations, scales, and appearances. The images were selected from different sensors and platforms, including Google Earth, the GF-2 satellite, and UAVs. Three versions of this dataset are available. The numbers of images and instances contained in the three versions of DOTA are summarized in Table 4. DOTA-V1.0 (Xia et al. 2018) and DOTA-V1.5 share the same images, which are split into training, validation, and test subsets. As an extension of DOTA-V1.0, DOTA-V1.5 annotates extremely small instances whose sizes are equal to or less than 10 pixels. Compared with the previous versions, DOTA-V2.0 (Ding et al. 2022) contains more images. In addition, many images were taken under an oblique view and a lower foreground ratio to approach real-world application scenes. The number of instances has increased to approximately 1.8 million. Moreover, it contains two test subsets, namely, test-dev and test-challenge. The latter comprises a greater number of object instances (approximately 1.1 million) and more complicated scenes, making the associated task more challenging.

*DIOR-R* (Cheng et al. 2022) is a large-scale dataset that contains 192,518 instances covering 20 common categories with notable interclass similarity and intraclass discrepancies.

**Table 3** Comparison of public RS image datasets

| | Dataset | Publication | Category | Quantity | Instance | GSD | Resolution |
|---|---|---|---|---|---|---|---|
| Early | SZTAKI-INRIA (Benedek et al. 2012) | TPAMI 2012 | 1 | 9 | 665 | – | $600 \times 500 \sim 1400 \times 800$ |
| | 3K vehicle (Liu and Mattyus 2015) | GRSL 2015 | 1 | 20 | 14,235 | 0.13m | $5516 \times 3744$ |
| | UCAS-AOD (Zhu et al. 2015) | ICIP 2015 | 2 | 2420 | 14,596 | – | $1280 \times 659$ |
| | VEDAI (Razakarivony and Jurie 2016) | JVCIR 2016 | 9 | 1210 | 3640 | 0.125m | $1024 \times 1024$ |
| | HRSC2016 (Liu et al. 2016) | GRSL 2016 | 25 | 1070 | 2976 | 0.4∼2 m | $300 \times 300 \sim 1500 \times 900$ |
| Modern | DOTA-V1.0 (Xia et al. 2018) | CVPR 2018 | 15 | 2806 | 188,282 | 0.1∼4.5m | $800 \times 800 \sim 20,000 \times 20,000$ |
| | DOTA-V1.5 | – | 16 | 2806 | 403,318 | 0.1∼4.5m | $800 \times 800 \sim 20,000 \times 20,000$ |
| | DOTA-V2.0 (Ding et al. 2022) | TPAMI 2022 | 18 | 11,268 | 1,793,658 | 0.1∼4.5m | $800 \times 800 \sim 29,200 \times 27,620$ |
| | FGSD (Chen et al. 2020) | arxiv 2020 | 43 | 5634 | 2612 | 0.12∼1.93m | $930 \times 930$ |
| | ShipRSImageNet (Zhang et al. 2021) | JSTAR 2021 | 50 | 3435 | 17,573 | 0.12∼6 m | $930 \times 930 \sim 1400 \times 1000$ |
| | DIOR-R (Cheng et al. 2022) | TGRS 2022 | 20 | 23,463 | 192,518 | 0.5∼30 m | $800 \times 800$ |
| | DroneVehicle (Sun et al. 2022) | TCSVT 2022 | 5 | 56,878 | 953,087 | – | $640 \times 512$ |
| | FAIR1M (Sun et al. 2022) | ISPRS 2022 | 37 | 42,796 | >1,000,000 | 0.3∼0.8m | $600 \times 600 \sim 10,000 \times 10,000$ |
| | GLH-Bridge (Li et al. 2024) | TPAMI2024 | 1 | 6,000 | 59,737 | 0.3∼1.0m | $2048 \times 2048 \sim 16,384 \times 16,384$ |

The previous version of DIOR-R, i.e., DIOR (Li et al. 2020), was initially released in 2019 using HBB annotations. Later, in 2021, OBB annotations were added to form the DIOR-R dataset. It includes 23,463 images chosen carefully from more than 80 countries, thereby possessing richer viewpoint, illumination, background, appearance, and occlusion variations. In particular, it contains some traffic infrastructures, such as train stations, expressway service areas, and airports, as well as some common categories in the suburbs, such as dams and wind mills, due to their significant value in transportation analyses. In addition, the GSD ranges from 0.5 m to 30 m, resulting in a large range of size variations. Thus, the rich diversity among its instances, images, and scales makes this dataset valuable for real-world tasks but presents challenges.

*Ship datasets* Recently, a series of ship datasets have drawn widespread attention due to the potential value of ship detection in fishing and maritime security scenarios. *HRSC2016* (Liu et al. 2016) is one of the most widely used datasets for evaluating oriented object detection algorithms. It covers more than 25 categories of ships with large scale, orientation, appearance, shape, and background (e.g., seas and ports) variations. *FGSD* (Chen et al. 2020) is a new fine-grained ship detection dataset that was expanded from HRSC2016. Its instances are classified into 43 categories, which are further divided into 4 high-level categories: submarines, aircraft carriers, civil ships, and warships. In addition to ships, a new category named docks is also annotated in this dataset for future research. *ShipRSImageNet* (Zhang et al. 2021) is the largest RS dataset for ship detection. It contains 3,435 images collected from xView (Lam et al. 2018), HRSC2016 (Liu et al. 2016), FGSD (Chen et al. 2020), the Airbus Ship Detection Challenge, and Chinese satellites. A total of 17,573 ships are divided into 50 categories. The dataset contains diverse spatial resolutions, scales, aspect ratios, backgrounds, and orientations.

*DroneVehicle* (Sun et al. 2022) is a large-scale RGB-infrared cross-modal vehicle detection dataset captured by UAVs. This dataset was released to address vehicle detection challenges encountered in smart city traffic management and disaster rescue scenarios, especially under conditions with insufficient lighting. The dataset includes two modalities, RGB images and infrared images, with an equal number of images in each modality, collectively forming image pairs. This dual-modality design can provide complementary information under different lighting conditions; e.g., RGB images provide rich color information, whereas infrared images excel in low-light conditions, as they are unaffected by darkness. In addition, the dataset covers a wide range of daytime and nighttime scenarios, including urban roads, rural areas, residential areas, and parking lots, ensuring the diversity and practicality of the data.

*FAIR1M* (Sun et al. 2022) is currently the largest fine-grained object detection dataset of high-resolution RS images, containing more than one million instances and over 40,000 images. All instances in this dataset were carefully annotated with OBBs, covering 5 main categories and 37 fine-grained subcategories, such as different types of aircrafts, ships, courts, roads, and vehicles. The images were sourced from different sensors and platforms, with target scenes covering hundreds of typical cities and towns as well as commonly used airports and ports globally, providing rich geographic information and practical application scenarios. The fine-grained annotations, intraclass and interclass variation similarities, large ranges of sizes and orientations, and complex scenes make this dataset extremely challenging while also promoting the development of object detection methods in the field of RS.

**Table 4** Comparison of the three versions of DOTA

| | | V1.0 | V1.5 | V2.0 |
|---|---|---|---|---|
| Images | Training | 1411 | | 1830 |
| | Validation | 458 | | 593 |
| | Test/Test-dev | 937 | | 2792 |
| | Test-challenge | – | | 6053 |
| | Total | 2806 | | 11,268 |
| Instances | Training | 98,990 | 210,631 | 268,627 |
| | Validation | 28,853 | 69,565 | 81,048 |
| | Test/Test-dev | 60,439 | 121,893 | 353,346 |
| | Test-challenge | – | – | 1,090,637 |
| | Total | 188,282 | 403,318 | 1,793,658 |

The number of images and instances of each split subset is counted

*GLH-Bridge* (Li et al. 2024) is a large-scale bridge detection dataset comprising 6,000 very-high-resolution (VHR) RS images sampled from diverse geographic locations around the globe. This dataset covers a wide range of scenarios and bridge types, enhancing its generalizability to real-world situations. Additionally, the various object scales and extreme aspect ratios contained in the dataset pose a formidable challenge for oriented object detection methods.

### 7.3 Discussion

The early datasets often featured limited numbers of instances and images, encompassing a narrow range of scenarios. Consequently, detection methods approached performance saturation on these datasets, making them unable to provide reliable evaluations. Modern datasets generally encompass more challenging and general scenarios with relatively complex backgrounds. They not only cover astonishing numbers of instances (up to millions) and fine-grained categories (e.g., FGSD and FAIR1M) but also possess large-scale images (up to $20,000 \times 20,000$ pixels), making them highly aligned with real-world application scenarios.

Both early and modern datasets play pivotal roles in propelling oriented object detection methods to new heights, making significant contributions to the field. HRSC2016 (Liu et al. 2016) is an early benchmark, but as detection methods reached their performance plateau on this dataset, it gradually fell out of favor. Although the subsequent datasets, e.g., FGSD (Chen et al. 2020) and ShipRSImageNet (Zhang et al. 2021), attempt to expand upon it, they failed to garner sufficient attention because of their relatively limited numbers of instances relative to those of larger-scale datasets. DOTA-V1.0 (Xia et al. 2018), as the most representative dataset for oriented object detection, serves as the most commonly used benchmark for evaluating the performance of detection methods. Subsequently, DOTA-V2.0 (Ding et al. 2022), which is characterized by its large scale and high level of challenge, and FAIR1M (Sun et al. 2022), which focuses on fine-grained oriented object detection, gradually became new benchmarks for evaluating the performance of cutting-edge methods.

Modern large-scale datasets provide solid data foundations for the deployment and implementation of real-world applications. By utilizing large-scale datasets for pretraining and transfer learning, the development and time costs of different methods can be significantly reduced, while the recognition accuracies of oriented object detection models can be improved, facilitating their application in various fields.

*City management* With the acceleration of urbanization and the continuous expansion of urban areas, traditional ground traffic monitoring systems have gradually revealed their limitations in terms of delayed responses. Due to their remarkable flexibility, UAVs are widely applied in traffic dispersion and traffic flow monitoring cases (Sun et al. 2022; Wang et al. 2022), emerging as a cutting-edge force in urban traffic management.

*Industrial inspection* Industrial facilities such as bridges and wind turbines often require significant amounts of labor and time for inspections, posing challenges for manual inspections (Cheng et al. 2022; Li et al. 2024). The incorporation of UAVs or satellites and intelligent detection technology can significantly improve the efficiency of inspections and ensure the safety of personnel.

*Port management* Through the use of advanced image processing algorithms and object detection technology, various objects within port areas, such as ships and port facilities, can be automatically identified from satellite imagery, thereby optimizing the efficiency of port operations and enhancing safety (Liu et al. 2016; Zhang et al. 2021).

*Security surveillance* The detection of critical objects in satellite imagery, such as airplanes and airports, plays a vital role in security surveillance scenarios (Ding et al. 2022).

## 8 State-of-the-art methods

As a comprehensive survey on oriented object detection, this paper introduces the recent advances and provides a structural taxonomy based on the existing detection frameworks, OBB representations, and other strategies in Sects. 3, 4, and 5, respectively. In this section, we select several publicly available detectors to compare them in a unified manner. Specifically, we take the DOTA-V1.0 dataset since it contains almost all the typical challenges related to this task, including arbitrary orientations, large-scale variations, and large aspect ratios. We report the performance achieved by the state-of-the-art detectors in terms of the mAP metrics and present the crucial modules of each detector in Table 5. According to the performance comparison and previous discussion, we concentrate on the key elements that have evolved in oriented object detection, including the detection frameworks, OBB regression techniques, feature representation approaches, and solutions to common issues.

(1) *Detection frameworks* Two-stage detectors achieve the best performance in terms of the mAP since they can extract accurate region-based features that are more suited for classification and regression tasks. The typical two-stage oriented object detection methods commonly design a rotated proposal generation scheme to obtain more accurate rotated proposals, such as the RoI Transformer (Ding et al. 2019) and Oriented RCNN (Xie et al. 2021). Similarly, the majority of one-stage and anchor-free detectors introduce a refined stage for aligning features, including $R^3$Det (Yang et al. 2021), $S^2$ANet (Han et al. 2022), CFA (Guo et al. 2021, 2022), and Oriented RepPoints (Li et al. 2022). Benefiting from their additional refined stage and advanced loss functions, one-stage detectors can also reach approximately equal accuracy to that of two-stage detectors. Despite achieving state-of-the-art performance in general object detection tasks, the DETR-based methods still lag behind the other competitors in oriented object detection, even with more training epochs. This may be because the query paradigm cannot adequately cover rotated objects. To this end, it is desirable to further investigate DETR-based methods that can compete in this field.

(2) *OBB regression techniques* Advanced loss functions are conducive to alleviating the problems caused by orientation parameters and achieving better regression effects, including the Gaussian distance-based loss (*e.g.,* GWD (Yang et al. 2021), KLD (Yang et al. 2021), and KFIoU (Yang et al. 2022)). These methods draw upon a trigonometric encoder and joint optimization to achieve strong performance. On the other hand, very large gaps are observed between different OBB representation methods; e.g., the midpoint offset representation scheme enables the Oriented RCNN to outperform the rotated Faster RCNN by approximately 2 mAP in the single-scale and multiscale results. In addition, the novel OBB representation methods elegantly avoid angular boundary discontinuities, thus enhancing their model performance, but they rely on complex post-processing operations and additional modules, including customized loss functions (e.g., the CIoU (Guo et al. 2021)) or assigners (e.g., APAA (Li et al. 2022)). Therefore, advanced loss functions and OBB representation methods are crucial for attaining improved regression accuracy.

(3) *Feature representation approaches* As some of the most important components in oriented object detection, backbone networks play a critical role in learning high-level semantic feature representations. The most widely used backbone networks include the ResNet (He et al. 2016; Xie et al. 2017) series and transformer architectures (Dosovitskiy et al. 2021; Xu et al. 2021; Zhang et al. 2023; Liu et al. 2021). Although transformer-based methods dominate the field of computer vision tasks, they significantly outperform their CNN-based counterparts in oriented object detection scenarios, achieving state-of-the-art performance. Specifically, Oriented RCNN-RVSA (Wang et al. 2022) and Oriented RCNN-STD (Yu et al. 2024), as the top-performing detectors based on transformers, outperform the Oriented RCNN (Xie et al. 2021) by $1.22\%$ and $2.22\%$ in terms of the mAP metric, respectively. Nevertheless, compared with CNNs, transformers suffer from longer training convergence times and expensive computing costs.

(4) *Solutions to common issues* Attention mechanisms and semantic mask modules, *e.g.*, SCRDet (Yang et al. 2019), TSH (Yu et al. 2023), and HRDet (Yao et al. 2024), are effective ways to reduce background noise and enhance object information. With respect to the issue of large aspect ratios, reweighted loss functions and novel assignment strategies can effectively improve the resulting detection accuracy, as seen in RPGAOD (Qiao et al. 2023), DFDet (Xie et al. 2024), CFL (Sun et al. 2024), and CGCDet (Wang et al. 2024). These components are removed during the inference process; thus, they do not affect the inference speed. On the other hand, as shown in Table 5, detectors with multiscale training and testing (MS),[4] achieve an average improvement of approximately $3\%$ in terms of the mAP, proving that MS is a useful strategy for alleviating scale variations. However, MS suffers from extremely long training and inference times, which are approximately 10 times greater than those of single-scale training and testing (SS). Overall, addressing common issues such as background noise, large aspect ratios, and scale variations is crucial for improving oriented object detection methods.

[4] MS generally first resizes the original images to three scales (i.e. $\{0.5, 1.0, 1.5\}$), which are then cropped to $1,024 \times 1,024$ patches with strides of 524. In contrast, SS crops only the original images into patches with sizes of $1,024 \times 1,024$ and strides of 824.

**Table 5** Comparisons of state-of-the-art methods on DOTA-V1.0

| Method | Publication | Baseline | Backbone | RPN | Assigner | Head | Reg loss | Cls loss | OBB representation | FE | mAP-SS | mAP-MS |
|---|---|---|---|---|---|---|---|---|---|---|---|---|
| Two-stage | | | | | | | | | | | | |
| Rotated faster RCNN (Ren et al. 2015) | NIPS 2015 | Faster RCNN | Res50 | – | – | – | – | – | – | | 73.40* | 78.49* |
| RoI transformer (Ding et al. 2019) | CVPR 2019 | Faster RCNN | Res50 | RRoI learner | – | – | – | – | – | | 75.63* | 80.43* |
| Oriented RCNN (Xie et al. 2021) | ICCV 2021 | Faster RCNN | Res50 | Oriented RPN | – | – | – | – | Midpoint Offset | | 75.69* | 80.02* |
| Gliding vertex (Xu et al. 2021) | TPAMI 2021 | Faster RCNN | Res50 | – | – | – | – | – | Gliding Vertex | | 75.02 | |
| ReDet (Han et al. 2021) | CVPR 2021 | RoI Transformer | ReRes50 | – | – | – | – | – | – | | 76.25 | 80.10 |
| OSKDet (Lu et al. 2022) | CVPR 2022 | Faster RCNN | Res50 | – | – | Anchor Free | – | GFL v2 | Unordered Keypoints | | 76.37 | 80.91 |
| AOPG (Cheng et al. 2022) | TGRS 2022 | Faster RCNN | Res50 | Rotated FCOS | – | – | – | – | – | | 75.22 | 80.66 |
| DEA (Liang et al. 2022) | TGRS 2022 | ReDet | ReRes50 | FCOS+RPN | – | – | – | – | – | | | 80.37 |
| RVSA (Wang et al. 2022) | TGRS 2022 | Oriented RCNN | ViT+RVSA | – | – | – | – | – | – | | 78.61 | 80.80 |
| RPGA-OD (Qiao et al. 2023) | TGRS 2023 | Oriented RCNN | Res50 | GRG-RPN | – | – | AAO | – | – | | 76.47 | 81.20 |
| QPDet (Yao et al. 2023) | TGRS 2023 | Oriented RCNN | Res50 | – | – | – | – | – | Quadrant Point | | 76.25 | 81.00 |
| FRIoU loss (Qian et al. 2023) | TGRS 2023 | Oriented RCNN | Res50 | – | – | – | FRIoU | – | – | | 76.45 | 80.78 |

**Table 5** (continued)

| Method | Publication | Baseline | Backbone | RPN | Assigner | Head | Reg loss | Cls loss | OBB representation | FE | mAP-SS | mAP-MS |
|---|---|---|---|---|---|---|---|---|---|---|---|---|
| CGC-Det (Wang et al. 2024) | TNNLS 2024 | RoI Transformer | Res50 | – | OCP-Guided | – | CGC | – | – | | 77.34 | 80.70 |
| STD (Yu et al. 2024) | AAAI 2024 | Oriented RCNN | ViT | – | – | MAEB | – | – | – | | | 82.24 |
| One-stage | | | | | | | | | | | | |
| Rotated RetinaNet (Lin et al. 2017) | ICCV 2017 | RetinaNet | Res50 | | – | – | – | – | – | | 68.80* | |
| SCRDet (Yang et al. 2019) | ICCV 2019 | RetinaNet | SF-Net | | – | – | IoU-Smooth L1 | – | – | MDA | 72.61 | |
| CSL (Yang and Yan 2020) | ECCV 2020 | RetinaNet | Res50 | | – | – | – | – | CSL | | 69.51* | |
| $R^3$Det (Yang et al. 2021) | AAAI 2021 | RetinaNet | Res50 | | – | FRM | – | – | – | | 70.18* | |
| DCL (Yang et al. 2021) | CVPR 2021 | $R^3$Det | Res50 | | – | – | – | – | DCL | | 71.21 | |
| DAL (Ming et al. 2021) | AAAI 2021 | RetinaNet | Res50 | | DAS | – | – | – | – | | 71.44 | |
| FoR-Det (Zhang et al. 2022) | TGRS 2021 | $R^2$SSD | Res50 | | – | – | FARL | | – | FRM | 71.44 | |
| GWD (Yang et al. 2021) | ICML 2021 | $R^3$Det | Res50 | | – | – | GWD | – | Gaussian | | 71.56 | |
| KLD (Yang et al. 2021) | NeurIPS 2021 | $R^3$Det | Res50 | | – | – | KLD | – | Gaussian | | 71.73 | |
| BCD (Yang et al. 2022) | TPAMI 2022 | $R^3$Det | Res50 | | – | – | BCD | – | Gaussian | | 72.22 | |

**Table 5** (continued)

| Method | Publication | Baseline | Backbone | RPN | Assigner | Head | Reg loss | Cls loss | OBB representation | FE | mAP-SS | mAP-MS |
|---|---|---|---|---|---|---|---|---|---|---|---|---|
| $S^2$ANet (Han et al. 2022) | TGRS 2022 | RetinaNet | Res50 | | – | FAM | – | – | – | | 74.12 | 79.42 |
| CFC-Net (Ming et al. 2022) | TGRS 2022 | RetinaNet | Res50 | | DAS | RARM | – | – | – | PAM | 73.50 | |
| DCFL (Xu et al. 2023) | CVPR 2023 | $S^2$ANet | Res50 | | DCFL | – | – | – | – | | 74.26 | |
| KFIoU (Yang et al. 2022) | ICLR 2023 | $R^3$Det | Res50 | | – | – | KFIoU | – | Gaussian | | 71.60 | |
| TCD (Zhang et al. 2023) | TGRS 2023 | RetinaNet | Res50 | | TCA | TCH | – | TCL | – | | 75.18 | 80.05 |
| FADL-Net (Fu et al. 2024) | TII 2024 | RetinaNet | Res50 | | GADL | – | – | JLRQ | – | | 74.80 | 79.97 |
| TIR-Net (Li et al. 2024) | TGRS 2024 | $S^2$ANet | Res50 | | – | SRK+RFR | – | – | – | | 75.23 | 80.63 |
| CFL (Sun et al. 2024) | TIM 2024 | $S^2$ANet | Res50 | | STS | CFS | – | – | – | | 75.35 | |
| Anchor-Free | | | | | | | | | | | | |
| DRN (Pan et al. 2020) | CVPR 2020 | CenterNet | H104 | | – | DRH | – | – | – | FSM | 70.70 | |
| O2-DNet (Wei et al. 2020) | ISPRS 2021 | FCOS | H104 | | – | – | – | – | Middle lines | | 72.80 | |
| CFA (Guo et al. 2021, 2022) | CVPR 2021 | RepPoints | Res101 | | – | CFA | CIoU | – | Convex-Hull | | 75.05 | |
| Oriented RepPoints (Li et al. 2022) | CVPR 2022 | RepPoints | Res50 | | APAA | – | – | – | Adaptive Points | | 75.97 | |
| GFL (Wang et al. 2022) | TGRS 2022 | CenterNet | Res50 | | – | – | GF-CSL | – | CSL | | 74.68 | 77.54 |

**Table 5** (continued)

| Method | Publication | Baseline | Backbone | RPN | Assigner | Head | Reg loss | Cls loss | OBB representation | FE | mAP-SS | mAP-MS |
|---|---|---|---|---|---|---|---|---|---|---|---|---|
| SASM (Hou et al. 2022) | AAAI 2022 | RepPoints | Res50 | | SA-S | – | SA-M | | – | | 74.92 | 77.19 |
| DHRec (Nie and Huang 2023) | TPAMI 2023 | FCOS | Res50 | | – | – | – | – | DHRec | | 74.57 | 78.97 |
| DRDet (Zhang et al. 2023) | TGRS 2023 | FCOS | Res50 | | – | OFE | – | – | DRLs | | 74.85 | 79.34 |
| TSH (Yu et al. 2023) | TGRS 2023 | FCOS | Res50 | | – | – | JQE | | PBL | ISM | 77.18 | |
| HRDet (Yao et al. 2024) | TCSVT 2024 | FCOS | Res50 | | – | – | EIoU | – | – | HMP | 74.11 | 78.90 |
| DFDet (Xie et al. 2024) | TGRS 2024 | FCOS | Res50 | | – | – | PIAS | – | – | CDMN | 74.71 | 80.37 |
| DETR-Based | | | | | | | | | | | | |
| AO2-DETR (Dai et al. 2022) | TCSVT 2023 | Deformable DETR | Res50 | | – | AOPR | – | – | – | | 70.91 | |
| ARS-DETR (Zeng et al. 2024) | TGRS 2024 | DINO | Res50 | | ARM | RDA | ARA-CSL | – | – | | 73.79 | |

(·)†* indicates the detection results provided by the MMRotate (Zhou et al. 2022). – indicates that the method follows the default modules of the corresponding baseline. *RPN* Region proposal networks. *Reg Loss* Regression loss. *Cls loss* Classification loss. *FE* Feature Enhancement Methods. *mAP-SS* Single-scale test performance on DOTA-V1.0. *mAP-MS* Multiscale test performance on DOTA-V1.0. *Res50* ResNet-50 (He et al. 2016). *ReRes50* Rotation Equivariant ResNet-50. *H104* Hourglass-104 (Newell et al. 2016). *ViT* Vision Transformer (Dosovitskiy et al. 2021)

## 9 Conclusions and future directions

Performing oriented object detection on RS images is an important and challenging task in the field of RS and has been actively investigated. As summarized in this survey, a variety of methods have rapidly developed in recent years, demonstrating remarkable progress. In this survey, we first review the evolution process from horizontal to oriented object detection and summarize the typical challenges. Next, we provide a structural taxonomy for the existing detection frameworks and highlight the milestone detectors. We also present a detailed elaboration of OBB regression techniques and feature representation approaches. Furthermore, we discuss the common issues encountered in RS scenarios and the corresponding methods for solving them. Finally, we summarize the commonly used datasets and compare the excellent methods that have emerged in recent years.

Despite the rising prominence of artificial intelligence, deep learning has rapidly advanced the development of oriented object detection methods. However, owing to the presence of complex and ever-changing real-world scenarios, the performance of deep learning still faces limitations, hindering robust and reliable practical applications. Over the years, a considerable amount of research effort has been dedicated to tackling the challenges of feature misalignment, spatial misalignment, and OBB regression, as well as common issues (e.g., complex backgrounds, scale variations, large aspect ratios, and the lack of annotated samples). These efforts have led to marked improvements in detection performance. Nevertheless, it is imperative to acknowledge that a considerable gap persists between the current detection capabilities and the demands of practical applications. More critically, several issues remain insufficiently addressed, posing significant barriers to achieving further advancements in oriented object detection technology.

*Low detection efficiency* Detection efficiency is a pivotal factor in real-world applications of detectors. The current state-of-the-art oriented object detection models were designed to be exceptionally complex to achieve superior detection accuracy. Nevertheless, their intricate network architectures markedly impede their detection efficiency, rendering them unsuitable for real-time applications.

*Imbalanced datasets* Modern datasets focus on general scenarios and contain images derived from a variety of different environments and contexts. While this diversity helps models learn a wider range of features, it may also lead to poor performance in specific scenarios (such as those with snow, fog, and occlusion). Additionally, common scenes or objects may constitute the majority of these datasets, whereas rare or special scenes or objects may be scarce. This data imbalance issue may cause models to develop biases toward certain scenes or objects during training, resulting in performance degradations.

*Detection in a single-modal image* The current research community is dedicated to exploring and developing oriented object detection methods for single-modal images. However, these methods are inherently constrained by their reliance on a solitary information source and the lack of contextual cues, resulting in heightened vulnerability to various interference sources, including variations in lighting conditions, occlusions, and shadows.

Given that the aforementioned issues have not yet been explored and studied in the oriented object detection field, we further share some insights concerning potential future research directions.

*Lightweight methods* The demand for real-time object detectors on resource-limited mobile devices is increasing, necessitating innovative solutions that can overcome hard-ware constraints. Thus, lightweight oriented object detection architectures are required to fulfill the requirements of mobile and embedded applications. A feasible approach for promoting the application of oriented object detection in real-world scenarios is to adopt meticulously designed efficient network architectures or leverage neural architecture searches (Xiong et al. 2021) to discover optimal architectures. These lightweight network architectures enhance the efficiency and accuracy of feature extraction while concurrently reducing model parameter count and computational burden. For example, several lightweight network structures, such as MobileNet (Howard et al. 2017) and ShuffleNet (Zhang et al. 2018), have gained widespread adoption in object detection tasks. Another feasible method is to perform model compressions to develop highly competitive, compact, and rapid detection models, including parameter pruning (Hanson and Pratt 1988; Han et al. 2015; Gao et al. 2024; Zhang et al. 2024), quantization (Song et al. 2016; Xu et al. 2023; Ding et al. 2024), and knowledge distillation models (Hinton et al. 2015; Zheng et al. 2023; Wang et al. 2024). These compression techniques have demonstrated marked effectiveness in bolstering the generalization capabilities of models and mitigating underfitting during the training processes of efficient object detection models.

*Mission-specific datasets* In light of the prevailing imbalances among the categories and scenarios contained within the current datasets, coupled with the emerging trend of research on multimodal large-scale models, we delineate the future dataset collection directions from three perspectives: scenario-specific datasets, multimodal datasets, and large-scale datasets.

Scenario-specific datasets can provide more refined and accurate data that are tailored to specific scenarios (e.g., severe weather conditions, or rare scenarios), thus empowering models to achieve superior performance within those specific scenarios. Moreover, these datasets can effectively address the scarcity of high-quality data in specialized scenarios, thereby enhancing the generalization capabilities of models and promoting the practical application of oriented object detection in specialized fields.

Compared with single-modal oriented object detection datasets, multimodal datasets integrate diverse data types that complement each other, providing a wealth of data resources that are essential for tackling intricate problems. By leveraging the connections and relationships among various data types and fusing information derived from multiple modalities, the accuracy and performance of models can be substantially enhanced, leading to more comprehensive and accurate analysis results. An interesting attempt is DroneVehicle (Sun et al. 2022), in which two modalities–RGB and infrared–offer complementary information across different lighting conditions. RGB images offer rich color information, whereas infrared images excel under low-light conditions and are not affected by darkness. In the future, with the continuous development of more efficient and accurate multimodal technologies (Chen et al. 2020; Radford et al. 2021; Li et al. 2022, 2023), multimodal datasets will play a pivotal role in a wide array of fields, driving continuous innovations in oriented object detection technology.

Owing to their exceptional representation and generalization capacities, large models have emerged as a focal point of the current research (Vaswani et al. 2017; Ho et al. 2020; Liu et al. 2021; Kirillov et al. 2023). By relying on extensive data samples, large models can discern intricate features and underlying patterns within the input data, thereby enhancing both their accuracy and generalizability. As the technology pertaining to large models continues to advance, their ability to handle complex scenarios and multimodal data will be further enhanced, which will increase the possibility of performing oriented object detec-tion. Consequently, there is a pressing need to develop high-quality, large-scale datasets that are tailored for oriented object detection. These datasets will not only provide a rich repository of samples but also serve as a robust validation platform to facilitate the training and evaluation of large models in this domain.

*Multimodal large models* Multimodal large models, as a critical avenue for the progression of artificial intelligence toward artificial general intelligence (AGI), have emerged as a focal point of the current research (Li et al. 2023; Kirillov et al. 2023).. However, the existing oriented object detection methods predominantly concentrate on extracting valuable information from single-modal images, neglecting the guidance provided by multimodal data. On the other hand, the scarcity of high-quality multimodal data is a fundamental bottleneck that impedes the advancement of multimodal large models within the domain of oriented object detection. A prominent direction for future research will involve the exploration and integration of multimodal data. This encompasses the amalgamation of text and visual large models; the incorporation of global positioning system (GPS), inertial measurement unit (IMU), and RS imagery; and the fusion of diverse sensors. By harnessing the synergies among these various data modalities, we can unlock new potential techniques for enhancing the accuracy and robustness of oriented object detection methods. As technology continues to evolve, driven by the proliferation of data and the expansion of application scenarios, multimodal large models are expected to significantly increase the performance achieved in oriented object detection tasks.

## Appendix A $\theta$-based representation

The $\theta$-based representation adopts a vector in the format of $(x, y, w, h, \theta)$ to define an OBB. The present approaches can be classified into two types according to their definitions of the angle $\theta$, including the OpenCV definition (which follows the OpenCV protocol) and the long edge definition. As shown in Fig. 16a, the former defines $\theta$ as the acute (or right) angle between the OBB and the $x$-axis, leading to $\theta \in (0, \frac{\pi}{2}]$. Note that the width $w$ is defined as the side of the acute angle and can be shorter than the height $h$, which is shown at the top of Fig. 16a. To address this issue, the long edge definition sets $\theta$ as the angle between the long edge of the OBB and the $x$-axis. Therefore, the angular range is $[-\frac{\pi}{2}, \frac{\pi}{2})$ (Ding et al. 2019; Han et al. 2021) or $[-\frac{\pi}{4}, \frac{3\pi}{4})$ (Han et al. 2022), which are shown in Fig. 16b and Fig. 16c, respectively. As shown at the bottom of Fig. 16, the parameters used for the same OBB have significant differences in different OBB representation schemes.

Built upon well-designed horizontal detectors, most oriented object detectors predict OBBs through regression. In the $\theta$-based OBB representation, given an anchor box denoted by $b_a = (x_a, y_a, w_a, h_a, \theta_a)$, the detectors first predicts the offsets between it and the predicted OBB:

$$\begin{aligned} t_x^p &= \frac{x_p - x_a}{w_a}, t_y^p = \frac{y_p - y_a}{h_a}, \\ t_w^p &= \log \frac{w_p}{w_a}, t_h^p = \log \frac{h_p}{h_a}, t_\theta^p = f\left(\frac{\theta_p - \theta_a}{\pi}\right) \end{aligned} \tag{A1}$$

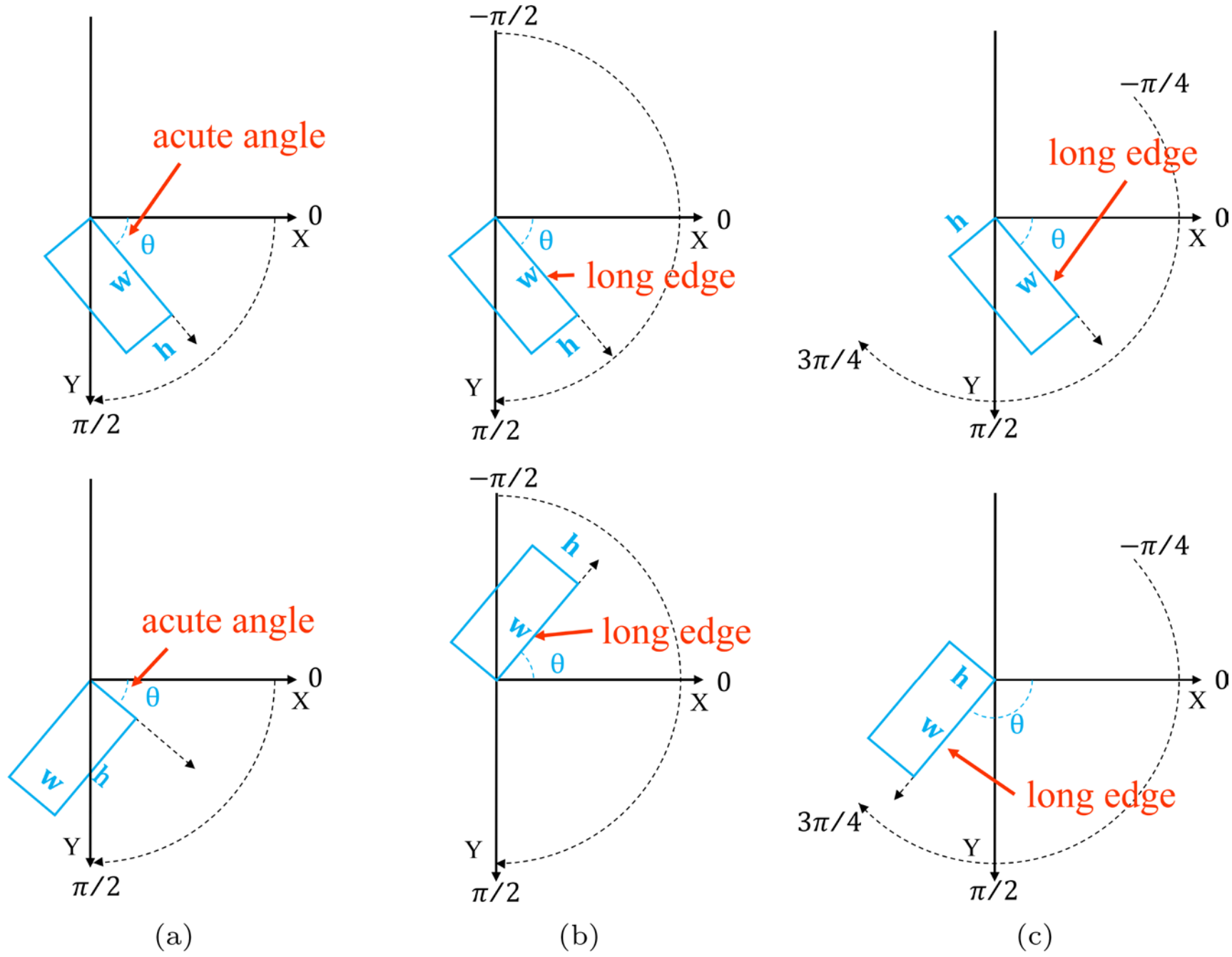

**Fig. 16** Definition of $\theta$-based representation. The OBBs depicted in the top or bottom row are the same. **a** OpenCV Definition ($\theta \in (0, \frac{\pi}{2}]$)(*Top* height is longer than width. *Bottom* width is longer than height). **b** Long edge definition with an angular range of $[-\frac{\pi}{2}, \frac{\pi}{2})$. **c** Long edge definition with an angular range of $[-\frac{\pi}{4}, \frac{3\pi}{4})$

where $b_p = (x_p, y_p, w_p, h_p, \theta_p)$ denotes the predicted OBB. $f(\cdot)$ is used to ensure that the angle difference remains within the preset range, thus avoiding the impact of the PoA. Similarly, the GT offsets are denoted by

$$
\begin{aligned}
t_x^g &= \frac{x_g - x_a}{w_a}, t_y^g = \frac{y_g - y_a}{h_a}, \\
t_w^g &= \log \frac{w_g}{w_a}, t_h^g = \log \frac{h_g}{h_a}, t_\theta^g = f\left(\frac{\theta_g - \theta_a}{\pi}\right)
\end{aligned}
\tag{A2}
$$

where $b_g = (x_g, y_g, w_g, h_g, \theta_g)$ denotes the GT OBB. Hence, the objective function for the regression task is as follows:

$$
L_{reg} = \sum_{i \in \{x,y,w,h,\theta\}} L_n(t_i^p - t_i^g) \tag{A3}
$$

where $L_n(\cdot)$ denotes the $L_n$ norm, and the smooth $L_1$ loss (Girshick 2015) is widely adopted. Owing to the PoA problem (Qian et al. 2021, 2022; Yang et al. 2021, 2022), the OBB regression process encounters the following challenges.

(1) *Metric-loss inconsistency* Although the majority of detectors adopt the smooth L1 loss as their objective function for regression, the most commonly used metric for local-ization is the RIoU. Therefore, an inconsistency is present between the loss function and the evaluation metric. This implies that an optimal choice for the regression task may not guarantee high localization accuracy in terms of the RIoU. Moreover, a good regression loss function should consider the central point distance, aspect ratio, and overlap area, which have been demonstrated to be effective in horizontal object detection tasks (Rezatofighi et al. 2019; Zheng et al. 2020). However, the aspect ratio and the overlap area can be easily disregarded by the smooth L1 loss.

We illustrate the metric-loss inconsistency in Fig. 17. As shown in Fig. 17a, the top and bottom rows have different angle differences, while the aspect ratios of the OBBs on the left are different from those on the right. Moreover, the center points, widths, and heights of the four cases are the same. The orange area denotes the IoU between a pair of OBBs. Note that the regression loss is sensitive to angle variances but remains unchanged under different aspect ratios. Specifically, when the aspect ratio varies, the union of two OBBs changes, but their intersection is constant, causing a change in the RIoU. The same conclusion can be drawn from Fig. 17b, which shows the variation curves exhibited by the RIoU and smooth L1 loss w.r.t. the aspect ratio under different angle differences. Note that the RIoU changes drastically, but the smooth L1 loss remains constant. Furthermore, Fig. 17c shows the variation curves exhibited by the RIoU and smooth L1 loss w.r.t. the angle under different aspect ratios. In the neighborhood of 0, both losses are consistent in their monotonicity but not in their convexity. The RIoU changes more intensely than the smooth L1 loss does when the angle difference is close to zero.

(2) *Angular boundary discontinuities and the square-like problem*

Because of the PoA problem (Yang et al. 2021, 2022; Qian et al. 2021, 2022), the smooth L1 loss suffers from the angular boundary discontinuities, which is illustrated in Fig. 18. Specifically, a small angle difference may cause a large loss change when the angular value approaches the angular boundary range. Figure 18a shows an ideal OBB representation, where the predicted and GT OBBs differ only slightly in terms of their angles and center points. For OBBs with the OpenCV definition, their angular value must be an acute or right angle, i.e., $\theta \in (0, \frac{\pi}{2}]$, as shown in Fig. 18b. As a result, the angle difference between the two OBBs increases sharply as the angular value approaches 0 or $\frac{\pi}{2}$. In addition, the width of the predicted OBB is the short edge, whereas the width of the GT OBB is the long edge, causing significant regression losses for the width and height. For OBBs under the long edge definition with an angular range of $[-\frac{\pi}{2}, \frac{\pi}{2})$, an angular boundary discontinuity leads to a significant angle difference, i.e., $|\theta_g - \theta_p| \approx \pi$, as shown in Fig. 18c. This problem also occurs in the long edge definition with an angular range of $[-\frac{\pi}{4}, \frac{3\pi}{4})$ when the angular value is close to $-\frac{\pi}{4}$ or $\frac{3\pi}{4}$.

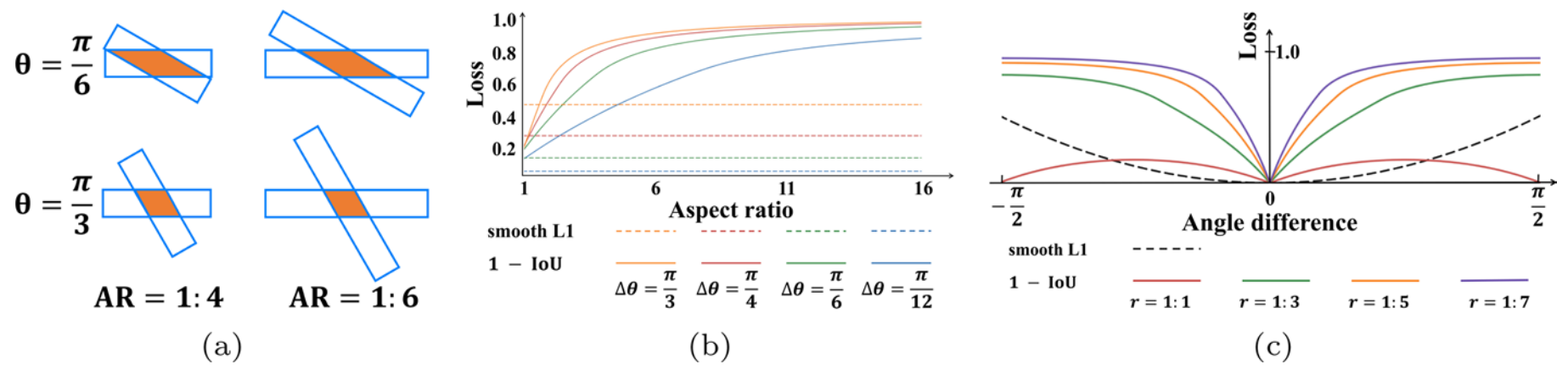

**Fig. 17** Comparison between metric and loss (Qian et al. 2021, 2022; Yang et al. 2021). **a** A sketch of RIoU change caused by angle and aspect ratio (AR) variation. **b** and **c** depict the changes of the regression loss and RIoU with aspect ratio and angle difference, respectively

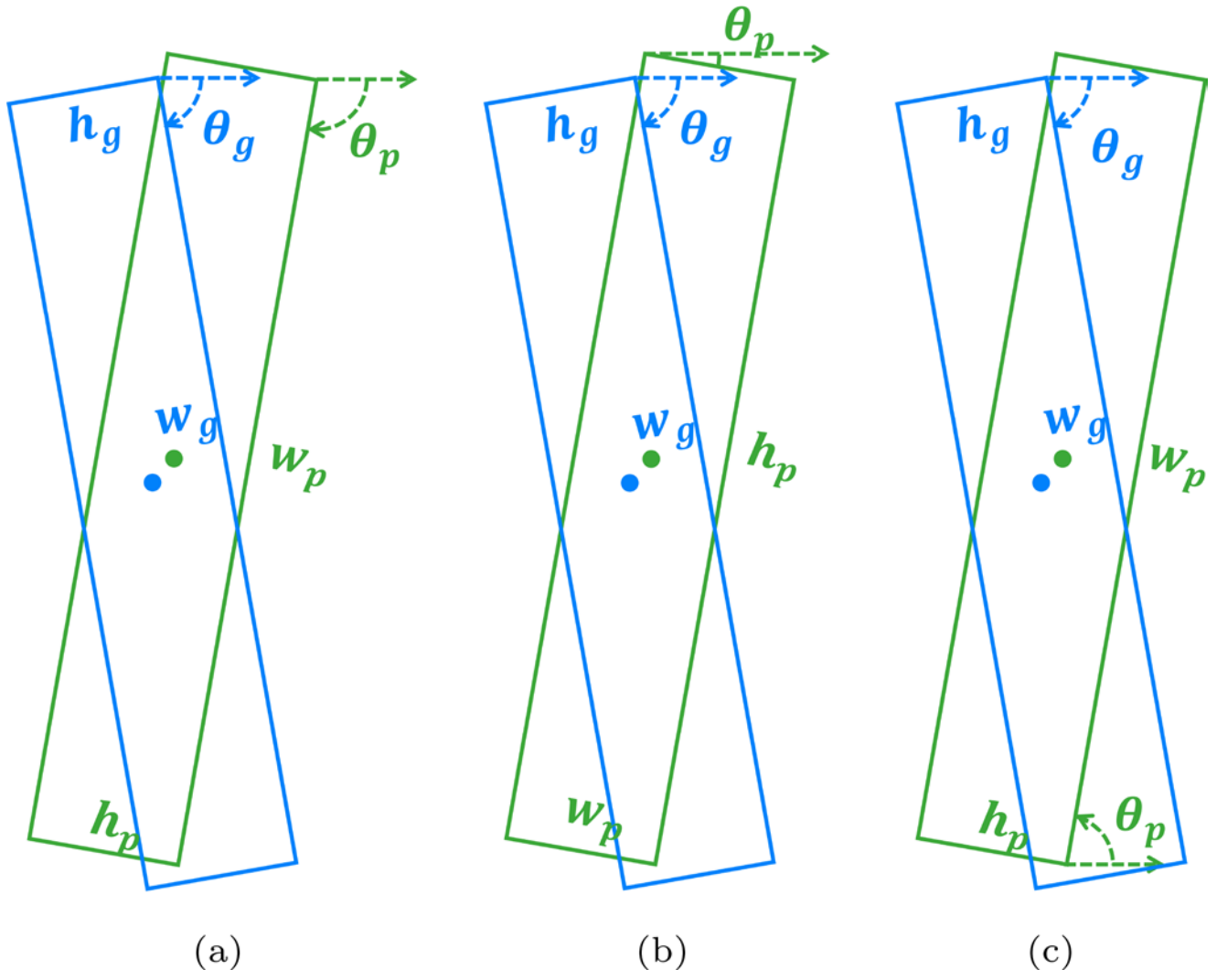

**Fig. 18** Illustration of angular boundary discontinuity (Yang et al. 2021). The predicted and GT OBB are represented by green and blue, respectively. **a** The ideal form of OBB representation. The two OBBs only differ slightly in terms of the angle and center point. **b** OBB representation with OpenCV definition, encountering PoA and exchangeability of edges (EoE). **c** OBB representation with long edge definition, encountering a significant angle difference

For square-like objects, including storage tanks and roundabouts, the long edge definition encounters a so-called square-like problem due to angle parameter differences (Yang et al. 2021, 2021, 2022). As shown in Fig. 19, when the aspect ratio is close to 1 but the length and width of the predicted OBB are opposite to those of the GT, the corresponding angle will differ by approximately $\frac{\pi}{2}$, leading to a large regression loss even if the RIoU is approximately 1.

## Appendix B Quadrilateral representation

The quadrilateral representation denotes an OBB as a vector $(x_1, y_1, x_2, y_2, x_3, y_3, x_4, y_4)$, where $(x_i, y_i)$ indicates the image coordinates of the $i_{th}$ vertex arranged in a clockwise order (Xu et al. 2021). This representation sch can compactly enclose oriented objects with large deformations and has been widely adopted for annotating objects in large-scale RS datasets, including DOTA (Xia et al. 2018; Ding et al. 2022) and HRSC2016 (Liu et al. 2016). The top-left vertex relative to the object orientation is chosen as the starting point $(x_1, y_1)$, as shown in Fig. 20a.

Under the quadrilateral representation, the detector outputs a vector $(\Delta x_1^p, \Delta y_1^p, \Delta x_2^p, \Delta y_2^p, \Delta x_3^p, \Delta y_3^p, \Delta x_4^p, \Delta y_4^p)$, where $(\Delta x_i^p, \Delta y_i^p)$ represent the relative offsets between the *i*-th vertex of the predicted OBB and the corresponding anchor box. Then, the predicted offsets are used to approximate the GT coordinate offsets $(\Delta x_1^g, \Delta y_1^g, \Delta x_2^g, \Delta y_2^g, \Delta x_3^g, \Delta y_3^g, \Delta x_4^g, \Delta y_4^g)$ between the *i*-th vertex of the GT OBB and that of the anchor box. The regression loss of the quadrilateral OBB representation can be expressed as follows:

$$L_{reg} = \sum_{i=1}^{4} \left[ L_n \left( \Delta x_i^p - \Delta x_i^g \right) + L_n \left( \Delta y_i^p - \Delta y_i^g \right) \right] \tag{B4}$$

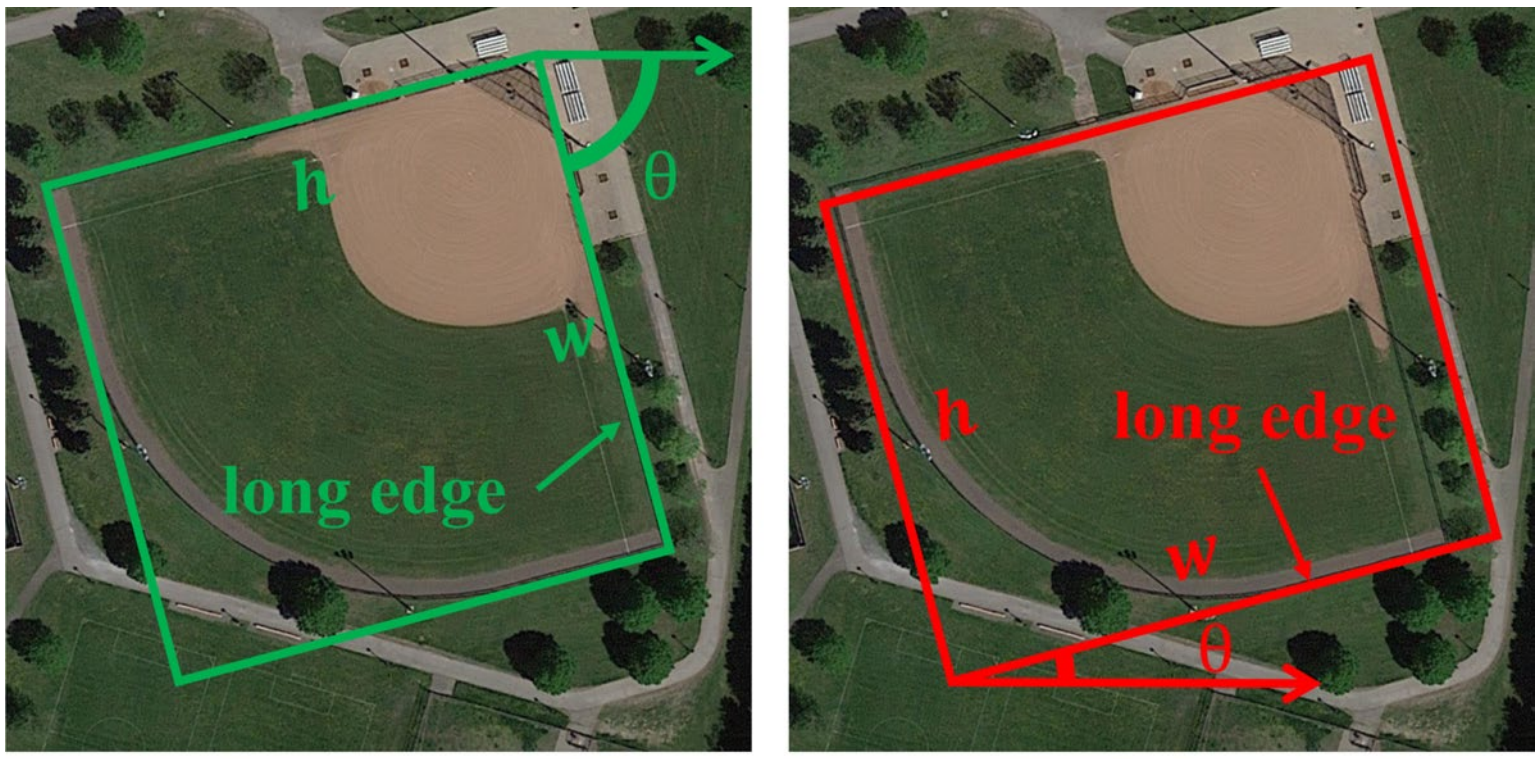

(a) GT OBB (b) Predicted OBB

**Fig. 19** Illustration of the square-like problem (Yang et al. 2021)

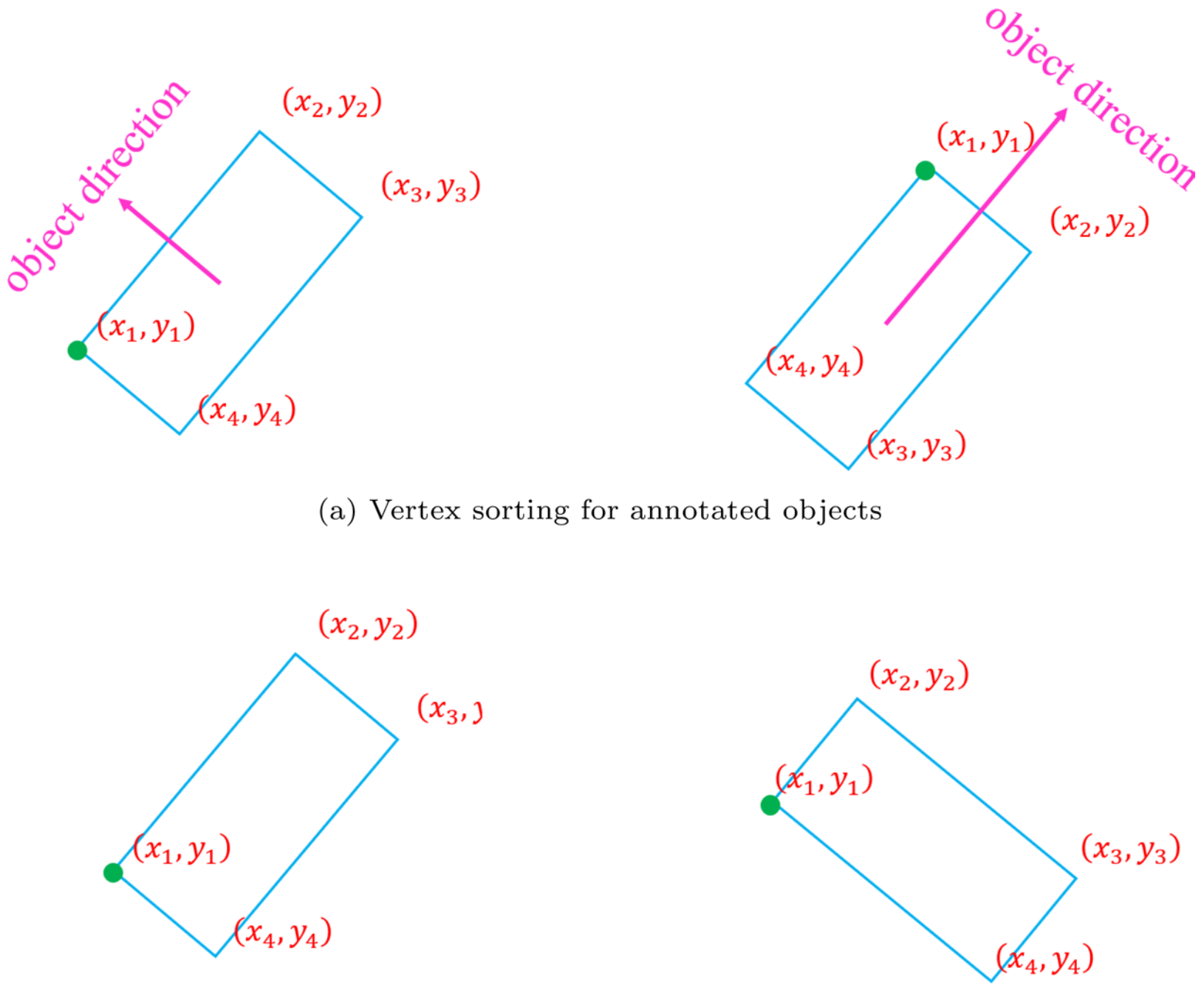

(b) Vertex sorting in detection process

**Fig. 20** Definition of quadrilateral representation. *Top* the top-left vertex relative to the object orientation is chosen as the start point. *Bottom* the leftmost vertex is chosen as the starting point

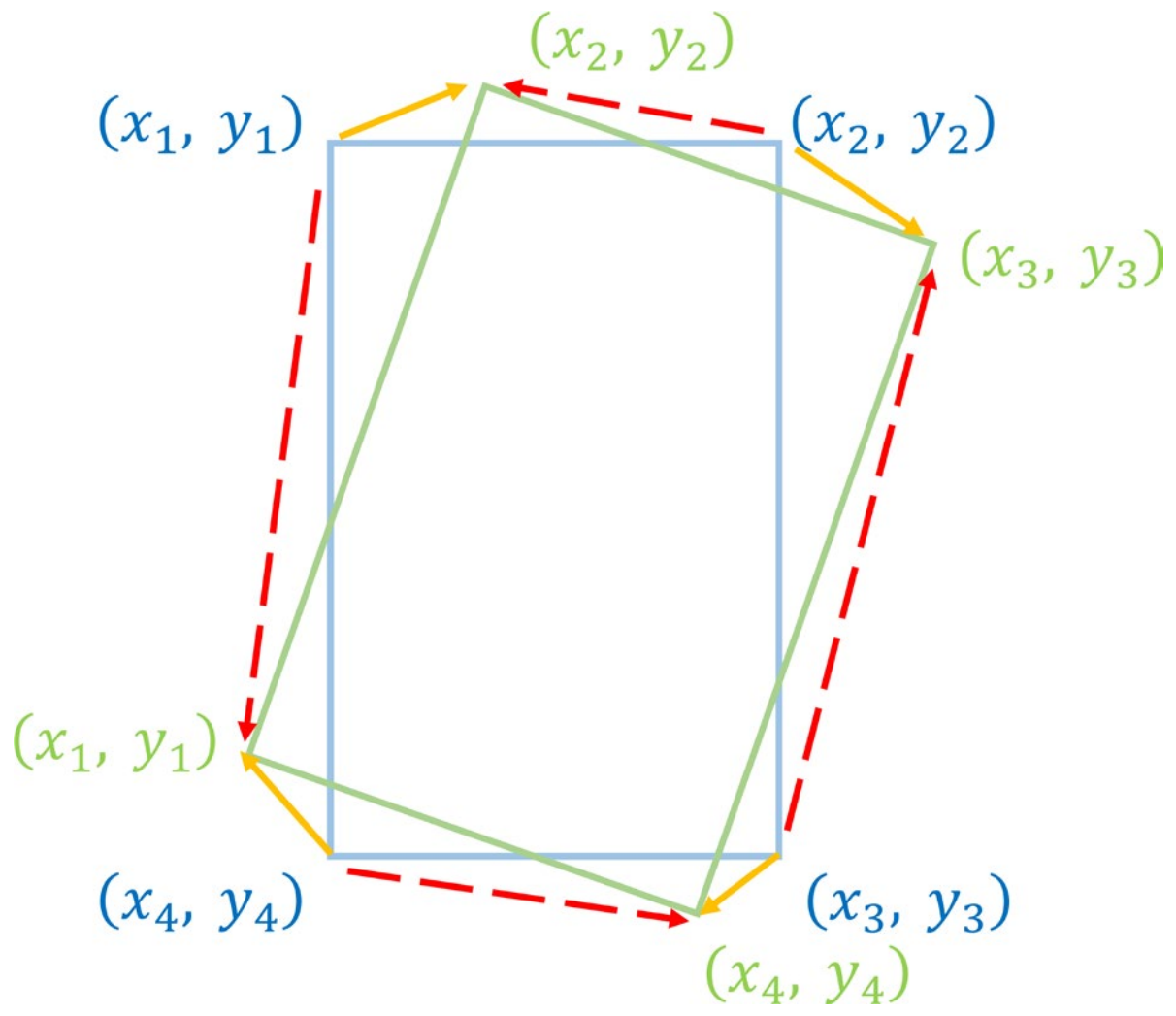

**Fig. 21** Illustration of vertexes sorting problem. The dashed line and solid line represent the actual and ideal regression forms, respectively

Generally, the anchor box selects the top-left vertex of the input image as the starting point. To ensure consistency, the leftmost vertices of the predicted OBB and the corresponding GT OBB are chosen as the starting points, as shown in Fig. 20b. However, an inappropriate vertex sorting process may cause inconsistencies between the vertex sequences of the anchor and the GT OBB, which is known as the vertex sorting or the corner sorting problem (Qian et al. 2021, 2022; Xu et al. 2021). Figure 21 shows a case of this problem. The anchor and the GT OBB are shown in blue and green, respectively, and the dashed and solid lines denote the actual and ideal vertex matchings during regression, respectively. In the ideal setting, the vertex matchings from the anchor to the GT are as follows:$(x_1, y_1) \rightarrow (x_2, y_2)$, $(x_2, y_2) \rightarrow (x_3, y_3)$, $(x_3, y_3) \rightarrow (x_4, y_4)$, and $(x_4, y_4) \rightarrow (x_1, y_1)$. However, in the actual regression case, the vertex matchings are $(x_1, y_1) \rightarrow (x_1, y_1)$, $(x_2, y_2) \rightarrow (x_2, y_2)$, $(x_3, y_3) \rightarrow (x_3, y_3)$, and $(x_4, y_4) \rightarrow (x_4, y_4)$. This inconsistency causes a large regression loss, confusing the network during the training process. Hence, determining the sequence of vertices in advance is critical for stabilizing the training process.

## Appendix C Datasets

In Table 3 of our main paper, we review a series of benchmark datasets for oriented object detection. However, space constraints prevent us from presenting all of them in detail. In this section, further details regarding the datasets mentioned in Sect. 7.2 are presented.

*SZTAKI-INRIA* (Benedek et al. 2012) contains 665 buildings in 9 multisensor aerial or satellite images taken from different cities. Owing to its small capacity, this dataset is used to evaluate traditional object detection algorithms.

*3K vehicle* (Liu and Mattyus 2015) was created for vehicle detection; it comprises 20 images and 14,235 vehicles. The images have resolutions of $5616 \times 3744$ and were captured by a DLR camera system at a height of 1,000 m above the ground. Therefore, the GSD is approximately 13 cm, leading to smaller scale variations. In addition, the images have similar backgrounds. Hence, this dataset is excluded from the algorithmic evaluation conducted on complicated scenes.

*VEDAI* (Razakarivony and Jurie 2016) was also proposed for vehicle detection, as it contains more categories and a wider variety of backgrounds, e.g., fields, grass, mountains, urban areas, etc., making the detection process more complicated. It comprises 1,210 images with resolutions of $1,024 \times 1,024$. The images were cropped from VHR satellite images with GSDs of 12.5 cm. However, the dataset consists of only 3,640 instances because images with too many dense vehicles are excluded. Notably, each image has four color channels, including three visible channels and one 8-bit near-infrared channel.

*UCAS-AOD* (Zhu et al. 2015) contains 7,482 planes in 1,000 images, 7,114 cars in 510 images, and 910 negative images. All the images in this dataset were cropped from Google Earth aerial images. In particular, the instances were carefully selected to ensure that their orientations were distributed evenly.

*DOTA* (Xia et al. 2018; Ding et al. 2022). Figure 22 shows the number of instances contained in each category of the training and validation subsets of DOTA-V1.0, V1.5, and V2.0. Note that the distributions of different categories are severely imbalanced. The instances of small vehicles and ships have large quantities. Nearly half of the other categories have quantities of less than 1,000, including planes, baseball diamonds, ground track fields, basketball courts, soccer ball fields, roundabouts, helicopters, container cranes, airports and helipads. Severe category imbalances cause models to severely overfit to many-shot categories but underfit to low-shot categories (Gupta et al. 2019; Cui et al. 2019; Wang

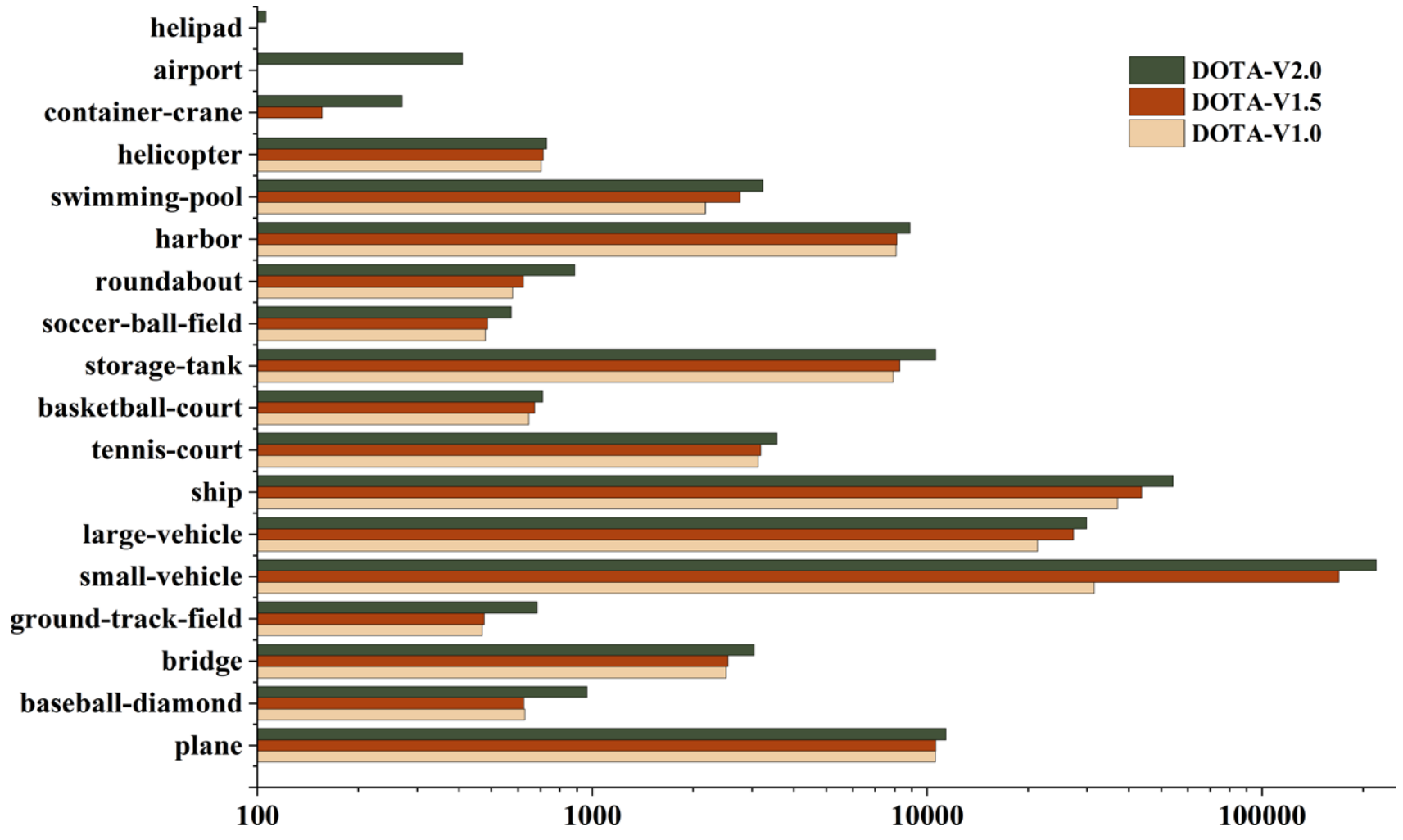

**Fig. 22** Number of instances for each category in training and validation subsets of DOTA-V1.0, V1.5, and V2.0 (Xia et al. 2018; Ding et al. 2022)

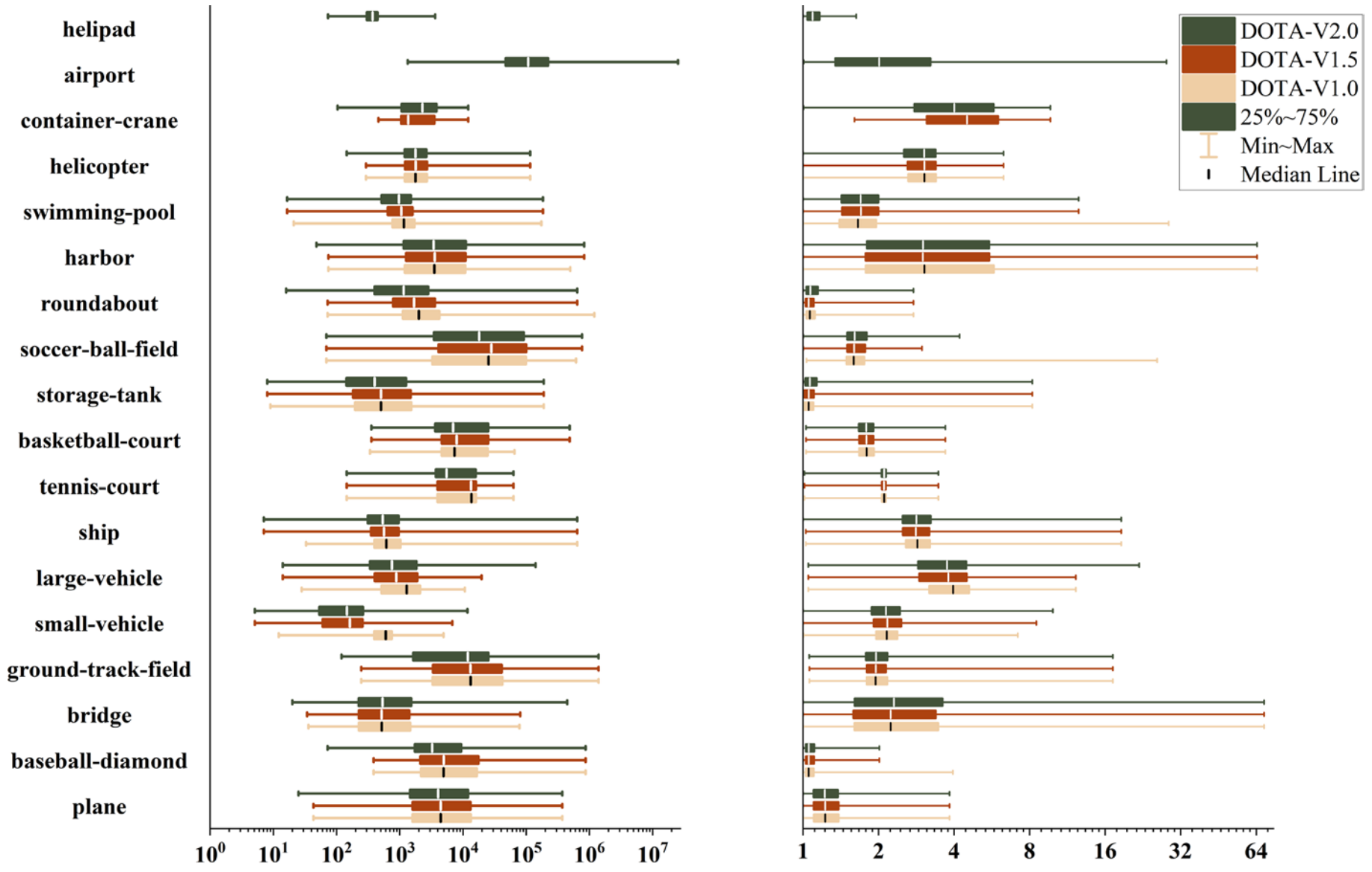

(a) Size distributions per category (b) Ratio distributions per category

**Fig. 23** Size and ratio distributions for each category in training and validation subsets of DOTA-V1.0, V1.5, and V2.0 (Xia et al. 2018; Ding et al. 2022)

et al. 2021). Figure 23 further summarizes the size and ratio distributions for each category in the three versions of DOTA. As shown in Fig. 23a, the minimum size is $3-4$ orders of magnitude lower than the maximum size for each category. Moreover, a large range of size differences are observed between various categories. Figure 23b indicates that the aspect ratios of different categories vary greatly. Furthermore, some categories, such as bridges, harbors, and airports, have extremely large aspect ratios. To date, DOTA has been the most challenging dataset for performing oriented object detection because of its large number of object instances, large aspect ratio, significant size variance, and complicated aerial scenes. All of these characteristics have made DOTA the de facto benchmark for evaluating the efficacy of oriented object detectors in recent years.

**Acknowledgements** This work was supported by the National Natural Science Foundation of China under Grant 12472189 and The Science and Technology Innovation Program of Hunan Province under Grant 2022RC1196.

**Author contributions** Kun Wang, Zi Wang, Zhang Li, and Erting Pan wrote the main manuscript text. Ang Su contributed to the data analysis, interpretation of results All authors reviewed the manuscript.

**Data availability** No datasets were generated or analysed during the current study.

## Declarations

**Competing interests** The authors declare no competing interests.

**Publisher's Note** Springer Nature remains neutral with regard to jurisdictional claims in published maps and institutional affiliations.

## Authors and Affiliations

**Kun Wang[1,2] · Zi Wang[1,2] · Zhang Li[1,2] · Ang Su[1,2] · Xichao Teng[1,2] · Erting Pan[1,2] · Minhao Liu[3] · Qifeng Yu[1,2]**

✉ Zhang Li
lizhang08@nudt.edu.cn

Kun Wang
wangkun21@nudt.edu.cn

Zi Wang
wangzi16@nudt.edu.cn

Ang Su
suang@nudt.edu.cn

Xichao Teng
tengari@buaa.edu.cn

Erting Pan
panerting@whu.edu.cn

Minhao Liu
lmh313@nudt.edu.cn

Qifeng Yu
yuqifeng@nudt.edu.cn

[1] College of Aerospace Science and Engineering, National University of Defense Technology, Deya Road, Changsha 410000, Hunan Province, China

[2] Hunan Provincial Key Laboratory of Image Measurement and Vision Navigation, National University of Defense Technology, Deya Road, Changsha 410000, Hunan Province, China

[3] Videogrammetry Innovation Center, Hunan Institute of Advanced Technology, Qingshan Road, Changsha 410000, Hunan Province, China